\documentclass{article} % For LaTeX2e
\usepackage{iclr2025_conference,times}

\usepackage{amsmath,amsfonts,bm}

\def\eqref#1{equation~\ref{#1}}
\def\1{\bm{1}}

\def\va{{\bm{a}}}

\def\vh{{\bm{h}}}

\def\vm{{\bm{m}}}

\def\vq{{\bm{q}}}

\def\vx{{\bm{x}}}
\def\vy{{\bm{y}}}
\def\vz{{\bm{z}}}

\def\mZ{{\bm{Z}}}

\DeclareMathAlphabet{\mathsfit}{\encodingdefault}{\sfdefault}{m}{sl}
\SetMathAlphabet{\mathsfit}{bold}{\encodingdefault}{\sfdefault}{bx}{n}

\newcommand{\E}{\mathbb{E}}

\newcommand{\KL}{D_{\mathrm{KL}}}

\DeclareMathOperator*{\argmax}{arg\,max}

\PassOptionsToPackage{xcdraw,table}{xcolor}
\usepackage{hyperref}
\usepackage{url}
\usepackage{amsmath}
\usepackage{graphicx}
\usepackage{amssymb}
\usepackage{longtable}
\usepackage{booktabs}
\usepackage{multirow}
\usepackage{makecell}
\usepackage{CJKutf8}
\usepackage{varwidth}
\usepackage{tikz}
\usepackage{mathrsfs}
\usepackage{ragged2e}
\usepackage{setspace}
\usepackage{enumitem}
\usepackage{pifont}
\usepackage{caption}
\usepackage{overpic}
\usepackage{color}
\usepackage{colortbl}
\usepackage[normalem]{ulem}
\usepackage{arydshln}
\usepackage{textcomp}
\usepackage{wasysym}
\usepackage{adjustbox}
\usepackage{kotex}
\usepackage{wrapfig}
\usepackage[ruled]{algorithm2e}

\definecolor{darkblue}{rgb}{0, 0, 0.7}
\hypersetup{colorlinks=true, citecolor=darkblue}

\title{Precise Localization of Memories: A Fine-grained Neuron-level Knowledge Editing Technique for LLMs}

\newcommand*{\affaddr}[1]{#1}
\newcommand*{\affmark}[1][*]{\textsuperscript{#1}}
\newcommand*{\email}[1]{\texttt{#1}}

\author{Haowen Pan\affmark[\textnormal{1}]~ Xiaozhi Wang\affmark[\textnormal{2}]~ Yixin Cao\affmark[\textnormal{3}]\thanks{Corresponding authors.}~ Zenglin Shi\affmark[\textnormal{4}]~ Xun Yang\affmark[\textnormal{1}]\footnotemark[1]~ Juanzi Li\affmark[\textnormal{2}]~ Meng Wang\affmark[\textnormal{4}] \\
\affaddr{\affmark[1]University of Science and Technology of China} \\
\affaddr{\affmark[2]Department of Computer Science and Technology, BNRist, Tsinghua University} \\
\affaddr{\affmark[3]School of Computer Science, Fudan University} ~ \affaddr{\affmark[4]Hefei University of Technology} \\
\email{phw1129@mail.ustc.edu.cn~ wangxz20@mails.tsinghua.edu.cn} \\
\email{caoyixin2011@gmail.com~ \{zenglin.shi,wangmeng\}@hfut.edu.cn} \\
\email{xyang21@ustc.edu.cn~ lijuanzi@tsinghua.edu.cn} \\
}

\iclrfinalcopy % Uncomment for camera-ready version, but NOT for submission.
\begin{document}

\maketitle
\begin{abstract}

Knowledge editing aims to update outdated information in Large Language Models (LLMs). A representative line of study is locate-then-edit methods, which typically employ causal tracing to identify the modules responsible for recalling factual knowledge about entities. However, we find these methods are often sensitive only to changes in the subject entity, leaving them less effective at adapting to changes in relations. This limitation results in poor editing locality, which can lead to the persistence of irrelevant or inaccurate facts, ultimately compromising the reliability of LLMs. We believe this issue arises from the insufficient precision of knowledge localization. To address this, we propose a \textbf{Fi}ne-grained \textbf{N}euron-level Knowledge \textbf{E}diting (\textbf{FiNE}) method that enhances editing locality without affecting overall success rates. By precisely identifying and modifying specific neurons within feed-forward networks, FiNE significantly improves knowledge localization and editing. Quantitative experiments demonstrate that FiNE efficiently achieves better overall performance compared to existing techniques, providing new insights into the localization and modification of knowledge within LLMs.\footnote{We release our code at~\url{https://github.com/opanhw/FiNE}.}
\end{abstract}

% \vspace{-8pt}
\section{Introduction}
% \vspace{-2pt}
\label{sec:introduction}

{Recently, various methods for the precise editing of outdated or wrong knowledge within Large Language Models (LLMs)~\citep{touvron2023llama, touvron2023llama2, jiang2024mixtral, dubey2024llama} have been proposed~\citep{mazzia2023survey, yao2023editing, wang2023knowledge}.} These methods include memory-based editors~\citep{mitchell2022memory, zheng2023can, hartvigsen2024aging, yu2024melo}, meta-learning approaches~\citep{de2021editing, mitchell2021fast, hase2023methods, han2023divide}, and locate-then-edit methods~\citep{dai2022kn, meng2022locating, meng2022mass, li2023pmet, gupta2024unified}. This paper primarily focuses on locate-then-edit methods, which have emerged as a promising and mainstream approach for knowledge editing in LLMs. A key representative of these approaches is ROME~\citep{meng2022locating}, which employs causal tracing to identify specific modules responsible for recalling facts about subject entities. The success of ROME has inspired subsequent methods, e.g., MEMIT~\citep{meng2022mass} and PMET~\citep{li2023pmet} that utilize causal tracing, establishing its role as a foundational technique in the field. Locate-then-edit methods offer critical insights into the precise storage locations of knowledge, enabling targeted modifications that enhance the reliability and accuracy of outputs from LLMs.
% These methods improve the accuracy of knowledge modifications and allow for a focused approach to specific pieces of information, which is essential for developing effective and reliable knowledge editing techniques.

\begin{figure}[t]
    \centering
    \vspace{-5pt}
    \includegraphics[width=0.95\linewidth]{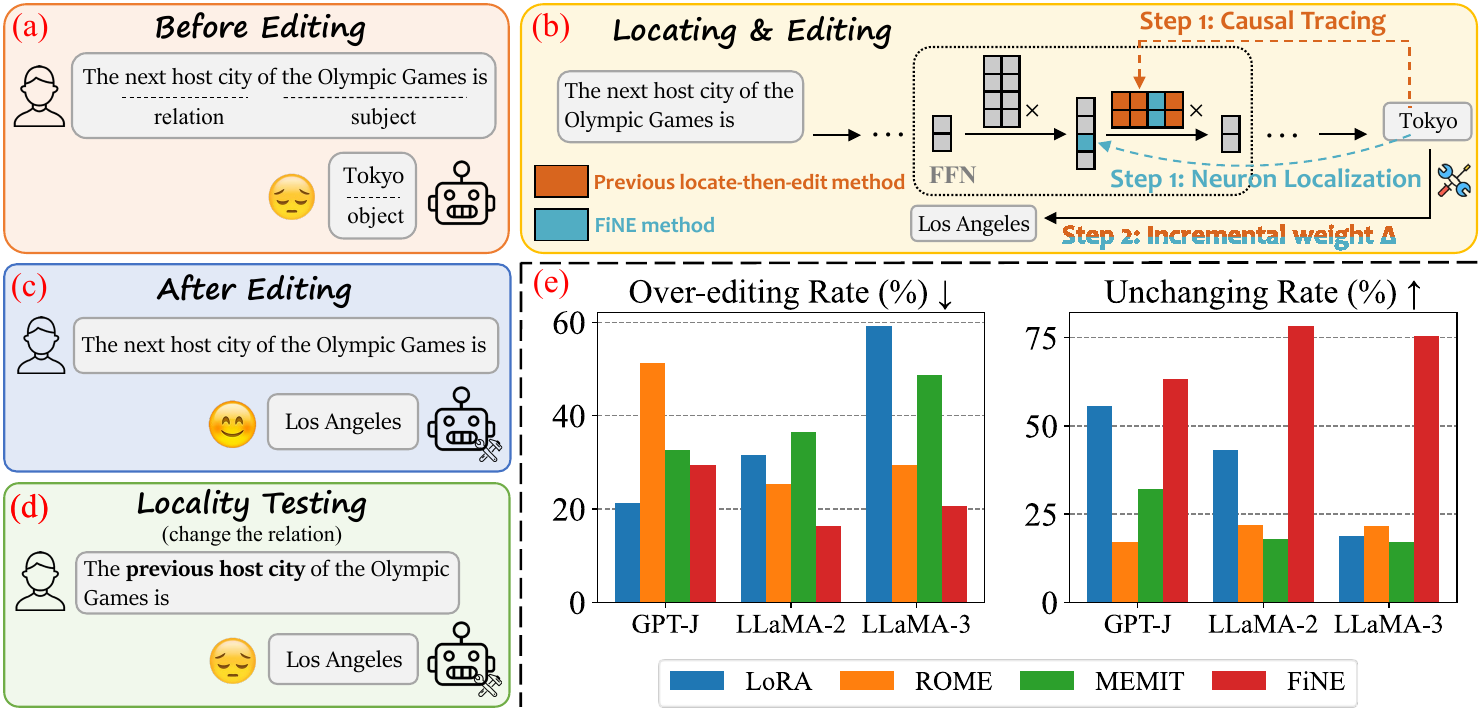} \\
    \caption{Previous locate-then-edit approaches (e.g., ROME and MEMIT) perform poorly in locality testing when changing the relation. \textbf{(a)} LLM makes a response ``Tokyo'' before knowledge editing. \textbf{(b)} We apply knowledge editing methods to edit the answer from ``Tokyo'' to ``Los Angeles''. \textbf{(c)} After editing, the model responses the target answer. \textbf{(d)} We then evaluate the post-edited model's locality and find that previous methods fail when changing the relation (i.e., outputting the target word even with unrelated inputs). \textbf{(e)} We also conduct quantitative experiments for the over-editing rate (lower values are better) and unchanging rate (higher values are better).}
    \label{fig:intro_example}
    \vspace{-5pt}
\end{figure}

However, \citet{hase2023does} question the validity of this localization method, noting that causal tracing offers limited insight into which Feed-Forward Network (FFN) layer should be edited to update existing knowledge. {The ineffectiveness of localization may cause the editing to be predominantly subject-driven.} One possible evidence is that locate-then-edit methods overly rely on the subject entity rather than the relation~\citep{wei2024does}. When we change the relation in locality testing, the post-edited model fails to produce correct answer and instead continues to generate the target object (see Figure~\ref{fig:intro_example}(d)). Furthermore, we conduct a pilot quantitative experiment on $\text {WikiData}_{counterfact}$ dataset in KnowEdit~\citep{zhang2024comprehensive} benchmark, and evaluate the over-editing rates and unchanging rates\footnote{See Appendix~\ref{appx:pilot_experiment_setup} for experimental setup details.}. As shown in Figure~\ref{fig:intro_example}(e), both ROME and MEMIT exhibit high over-editing rates and low unchanging rates, indicating causal tracing encounters issues during localization by focusing excessively on the subject and neglecting overall knowledge. Due to data construction issues with previously commonly used datasets, such as \textsc{CounterFact}~\citep{meng2022locating}, these problems have not been adequately exposed. These observations collectively indicate the localization of existing methods has significant flaws and lacks sufficient precision for guiding knowledge editing.

This motivates us to investigate more precise localization methods. Inspired by previous neuron-level analyses~\citep{dai2022kn, wang-wen-etal2022skill, schwettmann2023multimodal, pan2024finding}, we propose a \textbf{Fi}ne-grained \textbf{N}euron-level Knowledge \textbf{E}diting (\textbf{FiNE}) technique for a more precise localization of memories within LLMs. We first identify neurons in FFNs that are highly relevant to the knowledge to be edited and then update model weights at the locations of these neurons. Our neuron-level localization method provides a more finer-grained indication of the knowledge location compared to causal tracing and effectively avoids the problem of excessive focus on the subject. Furthermore, this approach benefits from fine-grained modifications to LLMs, resulting in a more efficient method that saves time and memory usage. Experiments on GPT-J~\citep{wang2021gpt}, LLaMA-2~\citep{touvron2023llama2}, and LLaMA-3~\citep{dubey2024llama} demonstrate that FiNE significantly outperforms existing locate-then-edit methods based on causal tracing, especially in locality metrics.

\vspace{-5pt}
\section{Related Work}
\vspace{-4pt}

\subsection{Model Editing Techniques}
\vspace{-2pt}
\textbf{Memory-based} For memory-based editors, some specific modules store the edit knowledge are used for post-edit response. SERAC~\citep{mitchell2022memory} stores edits in an explicit memory and learns to reason over them to modulate the base model’s predictions as needed. \citet{zheng2023can} explores in-context knowledge editing (IKE), a method without any gradient and parameter updating. GRACE~\citep{hartvigsen2024aging} is a lifelong model editing method that implements spot-fixes on streaming errors of a deployed model, ensuring minimal impact on unrelated inputs. Recently, \citet{yu2024melo} proposes MELO, a novel method that alters the behavior of LLMs by dynamically activating certain LoRA blocks according to the index built in an inner vector database.

\textbf{Meta-learning} Based on hypernetwork, several meta-learning methods have been proposed to edit models. \citet{de2021editing} presents KnowledgeEditor (KE), a method which can be used to edit knowledge and, fix bugs or unexpected predictions without the need for expensive re-training or fine-tuning. MEND~\citep{mitchell2021fast} introduces a collection of small auxiliary editing networks that use a single desired input-output pair to make fast, local edits to a pre-trained model's behavior. \citet{hase2023methods} proposes SLAG, a training objective for sequential, local, and generalizing updates with a better performance. \citet{han2023divide} proposes a novel divide-and-conquer framework, drawing on dynamic inference to break the zero-sum phenomenon in multiple edits.

\textbf{Locate-then-edit} Although prior research has explored knowledge storage mechanisms, the precise methods by which LLMs retain knowledge remain unclear. Studies have indicated that knowledge is often embedded within FFNs~\citep{geva2020transformer, geva2022transformer, dai2022kn}. Building on these, locate-then-edit methods have gained traction by first locating specific regions of knowledge storage and then executing targeted editing. A leading example is ROME, which innovatively employs causal tracing to pinpoint parameters intended for edits and directly updates them~\citep{meng2022locating}. This foundational work has paved the way for additional methods such as MEMIT~\citep{meng2022mass}, PMET~\citep{li2023pmet}, and EMMET~\citep{gupta2024unified}, enhancing the capacity to incorporate and modify larger quantities of knowledge. The advantages of locate-then-edit methods include increased precision in knowledge modification and the ability to selectively edit specific information, making them a vital advancement in the ongoing development of more effective and reliable LLMs.

\vspace{-2pt}
\subsection{Neuron Analyses in Transformer-based Models}
\vspace{-2pt}

Transformer~\citep{vaswani2017attention} is one of the most successful architectures designed for different tasks~\citep{song2024efficiently,li2023transformer} and there has been increasing interest in interpreting and analyzing the internal mechanisms of transformer-based models. Previous research has aimed to characterize the types of information encoded in individual neurons. \citet{dai2022kn} explores the identification of ``knowledge neurons'', which encode specific commonsense knowledge acquired during pre-training. Additionally, \citet{wang-wen-etal2022skill} presents a technique for identifying ``skill neurons'' in pre-trained transformer-based language models, which are crucial for specific tasks. \citet{schwettmann2023multimodal} explains how LLMs convert visual representations into corresponding texts by introducing a procedure for identifying ``multimodal neurons''. More recently, \citet{pan2024finding} proposes a novel method for finding ``multi-modal neurons'', which elucidates how multi-modal LLMs~\citep{zeng2024visual} bridge visual and textual concepts for captioning~\citep{song2024emotional}.

\vspace{-5pt}
\section{Preliminary}
\vspace{-2pt}
\label{sec:preliminary}

\textbf{Neurons in LLMs.} A decoder-only Transformer-based~\citep{vaswani2017attention} LLM (denoted as $\mathcal{M}$) typically consists of stacked self-attention and feed-forward layers. Each layer first performs multi-head self-attention and then applies a position-wise FFN. Residual connections and layer normalization are employed around each sub-layer. Following previous works~\citep{dai2022kn, wang-wen-etal2022skill, schwettmann2023multimodal, pan2024finding}, we investigate neurons within FFNs, as FFNs carry abundant information and knowledge. We denote the hidden states at layer $l$ as $\vh^l$, FFN output as $\vm^l$, and self-attention output as $\va^l$, respectively. The hidden states can be written as:
\begin{align}
\label{eq:pre_formula_1}
\vh^l &= \vh^{l-1} + \vm^l + \va^l, \\
\text{where}
\label{eq:pre_formula_2}
\quad\vm^l &= \mathbf{W}_{\text{out}}^{l}~\sigma\left(\mathbf{W}_{\text{in}}^{l} \gamma(\vx^l)\right),
\end{align}
$\vh^{0}$ is the embedding vector of input, $\sigma$ is an activation function, $\gamma$ is layernorm, $\mathbf{W}_{\text{in}}^{l}$ is the first linear layer and $\mathbf{W}_{\text{out}}^{l}$ is the second linear layer in the FFN, and $\vx^l$ represents the FFN input. For simplicity, let $\vq^l = \sigma\left(\mathbf{W}_{\text{in}}^{l} \gamma(\vx^l)\right)$. We regard $\vq^l_i$, the $i$-th element of $\vq^l$, as the activation of the $i$-th neuron on input $\vx^l$ at layer $l$. Each neuron in LLMs can be denoted as (L$l$.U$i$).

\textbf{Knowledge editing.} Extensive training on diverse datasets has endowed LLMs with a vast repository of knowledge~\citep{brown2020language, chowdhery2023palm, schott2023polyglot}. Formally, knowledge in LLMs can be denoted as triples like $(\text{subject}~s, \text{relation}~r, \text{object}~o)$~\citep{meng2022locating, meng2022mass}, such as $(s=\text{the Olympic Games}, r=\text{next host city}, o=\text{Tokyo})$. We define $p(\cdot)$ as a function that converts knowledge triples into prompt texts, for example, $p(\text{the Olympic Games}, \text{next host city})$ corresponds to ``The next host city of the Olympic Games is'' and $p(\text{the Olympic Games}, \text{next host city}, \text{Tokyo})$ corresponds to ``The next host city of the Olympic Games is Tokyo''. Let $(s, r, o^*)$ represents the updated knowledge. After editing, when the edited LLM (denoted as $\mathcal{M}'$) is given the input $p(s, r)$, it should return $o^*$ instead of $o$. For instance, if $o^*$ is ``Los Angeles'', the edited model should respond with ``The next host city of the Olympic Games is Los Angeles''.

\textbf{Evaluation for knowledge editing.} To effectively evaluate knowledge editing methods. \citet{zhang2024comprehensive} presents four essential criteria: Edit Success, Portability, Locality, and Fluency. \textbf{Edit Success} measures whether the post-edited model generates the expected output, which computes the accuracy of the outputs by $\E_{(s_j, r_j, o^*_j)\sim \mathcal{D}_{\text{edit}}} \1\{\argmax_y\mathbb{P}_{\mathcal{M}'}\left[y|p(s_j, r_j)\right]=o^*_j\}$. \textbf{Portability} evaluates how well the model can address the implications of an edit for real-world applications, which is computed by $\E_{(s_j, r_j, o^*_j)\sim \mathcal{D}_{\text{port}}} \1\{\argmax_y\mathbb{P}_{\mathcal{M}'}\left[y|p(s_j, r_j)\right]=o^*_j\}$. For example, when asked ``Is the next Olympic Games hosted in Tokyo?'' the post-edited model should answer ``No''. \textbf{Locality} examines whether an editing modifies the knowledge locally without influencing other unrelated knowledge, e.g., when asked, ``How often are the Olympic Games held?'' the model should still correctly respond with ``Every 4 years''. Locality can be calculated as $\E_{(s_j, r_j)\sim \mathcal{D}_{\text{loc}}} \1\{\argmax_y\mathbb{P}_{\mathcal{M}'}\left[y|p(s_j, r_j)\right]=\argmax_y\mathbb{P}_{\mathcal{M}}\left[y|p(s_j, r_j)\right]\}$. \textbf{Fluency} measures the model's generation ability by calculating a weighted average of bi-gram and tri-gram entropies~\citep{zhang2018generating}, denoted by $-\sum_k f(k) \log_2 f(k)$, where $f(·)$ is n-gram frequency distribution. A lower Fluency indicates a higher frequency of repeated words, signifying lower quality responses. These metrics offer a comprehensive assessment of methods' effectiveness, capturing various dimensions of performance and ensuring a robust analysis of the editing process.

\vspace{-5pt}
\section{Methodology}
\vspace{-2pt}

FiNE provides precise localization of knowledge within LLMs through a two-step process. In the first step, which differs significantly from causal tracing localization, it identifies key neurons in FFN layers that are closely associated with the knowledge to be edited. Subsequently, it updates model weights at these specific neuron locations.

\vspace{-2pt}
\subsection{Locating Neurons in LLMs}
% \vspace{-2pt}
\label{subsec:locating_neurons_in_llms}

Following previous work~\citep{dai2022kn, wang-wen-etal2022skill, schwettmann2023multimodal, pan2024finding} on selecting neurons in Transformer-based models, we present a neuron localization method for knowledge editing. Specifically, we hypothesize that a knowledge $(s_j, r_j, o_j)$ is stored in specific neurons, which are activated when LLMs receive input $(s_j, r_j)$, exhibiting a tendency to produce the output $o_j$. Therefore, our objective is to quantify contribution of each neuron to the current output and locate those neurons with higher impact. Following~\citet{pan2024finding}, who calculates contribution scores in multi-modal LLMs, we similarly compute contribution scores within LLMs.

For each token $t$ in the output $o_j$, we compute contribution score for each neuron $u_i$ at layer $l$ as:
\begin{align}
\label{eq:method_formula_1}
c_{(i, l, t)} = \vq^l_{i, -1}\cdot\left(\mathbf{W}_u\mathbf{W}_{\text{out}}^{l}\right)_{t, i},
\end{align}
where $\vq^l_{i, -1}$ is the activation output at the last token for neuron $u_i$ at layer $l$, $(\cdot)_{t, i}$ represents the $t$-th row and $i$-th column of the input matrix, and $\mathbf{W}_u$ is the unembedding matrix.

Here we regard $\mathbf{W}_u\mathbf{W}_{\text{out}}^{l}\in\mathbb{R}^{v\times d_m}$ as a projection function projecting from activations of the neurons to distribution of the vocabulary, where $d_m$ is the intermediate size and $v$ is the vocabulary size and regard $\vq^l_{i, -1}$ as a coefficient of the projection, respectively. This projection explicitly demonstrates the varying levels of focus that different neurons pay to different tokens, enabling us to calculate the contribution score. We provide detailed derivation in Appendix~\ref{appx:neuron_localization}.

After quantifying the contribution of each neuron, we rank all scores of neurons across all layers by the descending order and pick out top-$k$ neurons, denoted as $u^k$. These neurons are regarded as carriers of knowledge $(s_j, r_j, o_j)$ and make significant contributions to the output $o_j$. We follow the same procedure to locate neurons for each token $t$ in $o_j$, and use the set $\mathcal{U}_j=\left\{u^k_1, u^k_2, \cdots, u^k_{|o_j|}\right\}$ to represent neurons of $o_j$, where $|o_j|$ means the token length of $o_j$. {Algorithm~\ref{alg:neuron_localization} summarizes the entire process of neuron localization.}

\setlength{\columnsep}{15pt}

\begin{wrapfigure}[9]{r}{0.465\textwidth}
\vspace{-35pt}
\hspace{2em}
\RestyleAlgo{ruled}

\begin{spacing}{1.1}
\begin{algorithm}[H]
\fontsize{8pt}{8pt}\selectfont
\LinesNumbered

\caption{Neuron Localization}
\label{alg:neuron_localization}
\KwData{Knowledge $(s_j, r_j, o_j)$, LLM $\mathcal{M}$}
\KwResult{Neuron set $\mathcal{U}_j$ that carries knowledge $(s_j, r_j, o_j)$}
Initialize $\mathcal{U}_j=\emptyset$\;
 \For{each token $t$ in $o_j$}{
 Compute contribution of each neuron by Eqn.~\ref{eq:method_formula_1}\;
 $u^k \gets \text{select top-}k\text{~neurons by the descending order}$\;
 $\mathcal{U}_j \gets \mathcal{U}_j \cup \{u^k\}$\;
}
\end{algorithm}%
\end{spacing}
\end{wrapfigure}

\vspace*{-3pt}
\subsection{Updating Knowledge}
\label{subsec:updating_knowledge}
\vspace{-2pt}

We locate key neurons $\mathcal{U}_j$ of $o_j$ as described above, and then modify the model weights corresponding to locations of selected neurons to update the knowledge. For each neuron $u\in\mathcal{U}_j$, we assume that $u$ is the $i$-th neuron at layer $l$. Then we compute a vector $\vz\in \mathbb{R}^{d_h}$ and add it to the $i$-th row of matrix $\mathbf{W}_{\text{out}}^{l}$ for updating, where $d_h$ is the hidden size. If we stack vector $\vz$ for each neuron $u$ as $\mZ_j = [\vz_1~|~\vz_2~|~\cdots~|~\vz_{|\mathcal{U}_j|}]$, our objective can be succinctly represented as learning an optimized $\mZ_j$ based on $\mathcal{U}_j$, which is then applied to the model $\mathcal{M}$, resulting in a post-edited model $\mathcal{M}'$. Following \citet{meng2022locating}, the objective $\mathcal{L}(\mZ_j)$ consists of editing loss $\mathcal{L}_{\text{edit}}(\mZ_j)$, KL divergence $\mathcal{L}_{\text{KL}}(\mZ_j)$ and repetition penalty loss $\mathcal{L}_{\text{pen}}(\mZ_j)$. The editing loss utilizes negative log-likelihood to maximize the probability of the target $o^*_j$:
\vspace{-3pt}
\begin{align}
\label{eq:method_formula_2}
\mathcal{L}_{\text{edit}}(\mZ_j) = -\log\mathbb{P}_{\mathcal{M}'}\left[~o^{*}_j~|~p(s_j, r_j)~\right].
\end{align}
During the editing process, we aim to avoid altering unrelated knowledge or impacting the model's language capabilities. To this end, we add a KL divergence constraint of prompt that contains the subject and relation to the model, which is calculated by:
\vspace{-2pt}
\begin{align}
\label{eq:method_formula_3}
\mathcal{L}_{\text{KL}}(\mZ_j) = \KL \left(\mathbb{P}'_{\mathcal{M}'}\left[~y~|~p(s_j, r_j)~\right]~\Vert~\mathbb{P}'_{\mathcal{M}}\left[~y~|~p(s_j, r_j)~\right] \right),
\end{align}
where $\mathbb{P}'\left[\cdot\right]$ represents the probability distribution of output from position 1 to position $\ell_p-1$, assuming the length of the input prompt is $\ell_p$, which is different from $\mathbb{P}\left[\cdot\right]$.
Except KL divergence, to prevent the post-edited model from generating the editing target $o^*_j$ repeatedly, we also introduce a repetition penalty constraint. At the last position of the complete prompt $p(s_j, r_j, o^*_j)$, we use negative log-likelihood to maximize the probability of not generating $o^*_j$:
\vspace{-2pt}
\begin{align}
\label{eq:method_formula_4}
\mathcal{L}_{\text{pen}}(\mZ_j) = - \log\left(1 - \mathbb{P}_{\mathcal{M}'}\left[o^*_{j}~|~p(s_j, r_j, o^*_j)\right]\right).
\end{align}
Finally, we compute a weighted sum of editing loss, KL divergence and repetition penalty loss:
\vspace{-1pt}
\begin{align}
\label{eq:method_formula_5}
\mathcal{L}(\mZ_j) = \mathcal{L}_{\text{edit}}(\mZ_j) + \alpha\cdot\mathcal{L}_{\text{KL}}(\mZ_j) + \beta\cdot\mathcal{L}_{\text{pen}}(\mZ_j),
\end{align}
where $\alpha$ and $\beta$ are hyperparameters.

\vspace{-3pt}
\subsection{Layer Freezing}
\label{subsec:layer_freezing}
\vspace{-2pt}

In language models, the later layers are closely tied to the model's language capabilities~\citep{geva2020transformer, dai2022kn, wang-wen-etal2022skill, pan2024finding}. Arbitrary modifications to these later layers may impair model's linguistic abilities and result in responses with lower quality. To ensure the stability of LLMs, we implement layer freezing (LF) in our method. Specifically, for a LLM with $L$ layers, when locating neurons, we exclude the last $l_f$ layers, focusing only on the first $L-l_f$ layers. This ensures that no modifications are made to the last $l_f$ layers during the editing process.

\vspace{-6pt}
\section{Experiments}
\vspace{-4pt}

\subsection{Experimental Setup}
\label{subsec:experimental_setup}
\vspace{-2pt}

\textbf{Models and datasets.} We conduct experiments on the KnowEdit~\citep{zhang2024comprehensive} benchmark with GPT-J-6B (28 layers)~\citep{wang2021gpt}, LLaMA-2-7B (32 layers)~\citep{touvron2023llama2} and LLaMA-3-8B (32 layers)~\citep{dubey2024llama}. KnowEdit is an integrated benchmark for evaluating various knowledge editing methods, which contains six datasets for different evaluation types. We select three datasets including knowledge insertion and knowledge modification in our experiments: $\text {WikiData}_{counterfact}$~\citep{cohen2024evaluating}, $\text {WikiData}_{recent}$~\citep{cohen2024evaluating} and ZsRE~\citep{levy2017zero}. Notably, the locality evaluation in KnowEdit primarily focuses on changing the relation. The proportion of prompts where the subject changes in datasets $\text {WikiData}_{counterfact}$, $\text {WikiData}_{recent}$ and ZsRE is only 0.9\%, 0.1\%, and 0.0\%, respectively.

\textbf{Baselines.} We categorize baseline methods into three types of knowledge editing. The first category consists of methods that directly modify model parameters, such as Fine-Tuning (\textbf{FT}) and \textbf{LoRA}~\citep{wu2023eva}. The second category includes memory-based methods, for which we select In-context Knowledge Editing (\textbf{IKE})~\citep{zheng2023can}, which retrieves the most pertinent demonstrations. The third category focuses on locate-then-edit methods, which are central to our study. Although Knowledge Neurons (\textbf{KN}) is also a neuron-level knowledge localization method, it employs a significantly different technique than ours, selecting neurons via gradient-based attributions and modifying the corresponding FFN weights by adding scaled embedding vectors. Importantly, \textbf{ROME}~\citep{meng2022locating}, as a pioneer of causal tracing localization, has further advanced the locate-then-edit methods and significantly influenced the field, while \textbf{MEMIT}~\citep{meng2022mass} has built upon this foundation with notable enhancements. \textbf{PMET}~\citep{li2023pmet} serves as an improvement over MEMIT. Both ROME and MEMIT not only represent critical developments but have also achieved substantial popularity, making them essential comparisons in our work.

\textbf{Evaluation metrics.} As described in~\S~\ref{sec:preliminary}, we adopt four evaluation metrics in our experiments: Edit Success, Portability, Locality, and Fluency~\citep{zhang2024comprehensive}. \textbf{Portability} contains three parts: \textit{Subject Aliasing Accuracy (SAA)}, \textit{Logical Generalization Accuracy (LGA)} and \textit{Reasoning Accuracy (RA)}. Subject aliasing replaces the question’s subject with an alias or synonym to evaluate performance on other descriptions of the subject. Logical generalizations are changes that are semantically related to the modified fact and expected to change by the edit. Reasoning examines the reasoning ability with changed facts. \textbf{Locality} consists of two parts: \textit{Forgetfulness Accuracy (FA)} and \textit{Relation Specificity Accuracy (RSA)}. Forgetfulness evaluates whether the post-edited model retains the original objects in one-to-many relationships, whereas relation specificity evaluates whether any other attributes of the subject, which have been previously updated remain unaltered.

\vspace{-3pt}
\subsection{Quantitative Results}
\vspace{-2pt}

% ``KN + FiNE'' represents applying FiNE to edit the neurons localized by KN.
\begin{table}[t]
\captionof{table}{Editing results on $\text {WikiData}_{counterfact}$. 95\% confidence intervals are in parentheses.  \textbf{\textcolor[rgb]{0, 0.6, 0}{Green}} numbers indicate the best performance among locate-then-edit methods. \textbf{\textcolor[rgb]{0.7, 0.7, 0.7}{Grey}} numbers indicate invalid results\protect\footnotemark. Numbers with \uline{underline} indicate columnwise maxima for each model.}
\vspace{-3pt}
\label{tab:main_results}
\fontsize{7.4pt}{7.6pt}\selectfont
\centering
\begin{tabular}{l@{\hspace{8pt}}rrrrrrl}
\toprule
\multirow{2}{*}{\textbf{Method}} & \multirow{2}{*}{\textbf{Edit Succ.} $\uparrow$} & \multicolumn{3}{c}{\textbf{Portability} $\uparrow$} & \multicolumn{2}{c}{\textbf{Locality} $\uparrow$} & \multirow{2}{*}{\textbf{Fluency} $\uparrow$}  \\
\cmidrule(lr){3-5}
\cmidrule(lr){6-7}
& & \scriptsize\textbf{SAA} & \scriptsize\textbf{LGA} & \scriptsize\textbf{RA} & \scriptsize\textbf{RSA} & \scriptsize\textbf{FA} & \\
\midrule
\textit{GPT-J} & 21.5 (1.5) & 21.7 (1.6) & 14.8 (2.4) & 18.6 (1.6) & - & - & \uline{612.3 (3.1)} \\
\midrule
FT & 64.2 (1.6) & 47.3 (2.0) & 7.1 (1.9) & 21.3 (2.9) & 4.4 (0.6) & 6.4 (1.3) & 304.1 (7.6) \\
IKE & \uline{100.0 (0.0)} & \uline{98.0 (0.8)} & \uline{59.0 (6.1)} & \uline{61.5 (4.3)} & 60.6 (1.3) & 52.3 (3.1) & - \\
LoRA & \uline{100.0 (0.0)} & 75.2 (1.9) & 22.2 (3.1) & 40.3 (2.8) & 25.7 (1.6) & 51.4 (2.8) & 595.8 (4.1) \\
\midrule
KN & 18.1 (2.4) & 17.9 (2.4) & 10.8 (2.6) & 18.5 (2.2) & \bf \textcolor[rgb]{0.7, 0.7, 0.7}{80.2 (1.3)} & \bf \textcolor[rgb]{0.7, 0.7, 0.7}{80.6 (1.5)} & 580.0 (3.8) \\
% {KN + FiNE} & {66.6 (1.7)} & {48.2 (2.3)} & {14.3 (2.8)} & {24.2 (2.2)} & {76.8 (1.2)} & {\uline{\bf \textcolor[rgb]{0, 0.6, 0}{63.5 (2.4)}}} & {584.8 (3.5)} \\
ROME & 99.2 (0.5) & 74.1 (2.2) & 16.1 (2.6) & 29.2 (2.4) & 37.4 (1.3) & 33.1 (2.6) & 600.0 (3.6) \\
MEMIT & 99.5 (0.5) & 56.5 (2.5) & 16.7 (2.6) & 25.9 (2.1) & 53.2 (1.4) & 40.7 (2.8) & 591.6 (4.3) \\
PMET & 95.3 (0.9) & 54.1 (2.6) & 16.6 (2.6) & 25.3 (2.1) & 47.6 (1.5) & 36.8 (2.8) & \bf \textcolor[rgb]{0, 0.6, 0}{600.3 (3.6)} \\
\rowcolor[gray]{0.9} FiNE & \bf \textcolor[rgb]{0, 0.6, 0}{99.8 (0.1)} & \bf \textcolor[rgb]{0, 0.6, 0}{90.6 (1.4)} & \bf \textcolor[rgb]{0, 0.6, 0}{17.5 (2.7)} & \bf \textcolor[rgb]{0, 0.6, 0}{37.4 (3.5)} & \uline{\bf \textcolor[rgb]{0, 0.6, 0}{84.2 (1.1)}} & 54.2 (2.7) & 545.7 (7.3) \\
\midrule
\midrule
\textit{LLaMA-2} & 27.0 (1.5) & 27.8 (1.7) & 26.1 (2.9) & 26.2 (1.9) & - & - & \uline{583.3 (2.7)} \\
\midrule
FT & 47.3 (1.8) & 44.2 (1.9) & 17.9 (2.3) & 28.8 (2.0) & 59.5 (1.3) & 40.2 (2.7) & 500.9 (6.8) \\
IKE & \uline{100.0 (0.0)} & \uline{99.1 (0.5)} & \uline{70.2 (5.1)} & \uline{71.2 (3.8)} & 73.6 (1.1) & \uline{72.9 (2.5)} & - \\
LoRA & \uline{100.0 (0.0)} & 93.9 (1.0) & 29.9 (3.1) & 44.4 (3.1) & 73.5 (1.2) & 50.0 (2.7) & 559.3 (5.1) \\
\midrule
KN & 21.3 (2.3) & 21.8 (2.9) & 16.9 (2.7) & 24.6 (2.9) & \bf \textcolor[rgb]{0.7, 0.7, 0.7}{73.7 (2.1)} & \bf \textcolor[rgb]{0.7, 0.7, 0.7}{68.7 (3.5)} & 561.4 (6.3) \\
% {KN + FiNE} & {84.6 (1.5)} & {77.1 (1.9)} & {23.2 (3.0)} & {36.6 (3.2)} & {59.4 (1.4)} & {40.6 (2.9)} & {447.3 (10.2)} \\
ROME & 98.7 (0.6) & 72.2 (2.2) & 25.8 (2.8) & 35.1 (2.4) & 49.1 (1.2) & 40.5 (2.7) & \bf \textcolor[rgb]{0, 0.6, 0}{577.3 (3.3)} \\
MEMIT & 98.0 (0.7) & 76.2 (2.1) & 25.2 (2.8) & 35.0 (2.5) & 45.0 (1.3) & 40.1 (2.8) & 561.9 (4.6) \\
PMET & 94.8 (1.0) & 56.7 (2.5) & 27.2 (3.0) & 34.9 (2.4) & 64.5 (1.4) & 50.0 (2.8) & 576.1 (3.4) \\
\rowcolor[gray]{0.9} FiNE & \bf \textcolor[rgb]{0, 0.6, 0}{99.9 (0.2)} & \bf \textcolor[rgb]{0, 0.6, 0}{89.8 (1.4)} & \bf \textcolor[rgb]{0, 0.6, 0}{28.8 (3.0)} & \bf \textcolor[rgb]{0, 0.6, 0}{41.5 (3.0)} & \uline{\bf \textcolor[rgb]{0, 0.6, 0}{92.6 (1.0)}} & {\bf \textcolor[rgb]{0, 0.6, 0}{65.0 (2.8)}} & 542.3 (5.1) \\
\midrule
\midrule
\textit{LLaMA-3} & 23.1 (1.5) & 23.1 (1.7) & 21.7 (3.0) & 22.8 (1.9) & - & - & \uline{607.1 (2.9)} \\
\midrule
FT & 44.6 (1.9) & 45.0 (2.0) & 8.4 (1.7) & 23.9 (2.2) & 28.7 (1.3) & 14.2 (2.0) & 351.7 (9.8) \\
IKE & 61.8 (1.5) & 60.3 (1.8) & \uline{41.3 (5.2)} & {38.6 (3.3)} & 67.7 (1.2) & \uline{65.7 (2.6)} & - \\
LoRA & \uline{100.0 (0.0)} & 79.5 (1.8) & 23.2 (2.9) & \uline{45.6 (3.3)} & 17.5 (1.2) & 29.8 (2.5) & 455.7 (11.2) \\
\midrule
KN & 17.1 (2.1) & 18.1 (2.7) & 14.9 (2.6) & 19.2 (2.1) & \bf \textcolor[rgb]{0.7, 0.7, 0.7}{82.6 (1.6)} & \bf \textcolor[rgb]{0.7, 0.7, 0.7}{87.6 (2.3)} & 593.7 (6.8) \\
% {KN + FiNE} & {61.9 (1.8)} & {55.8 (2.5)} & {14.5 (2.6)} & {34.0 (2.3)} & {84.0 (1.5)} & {56.7 (2.5)} & {546.9 (7.3)} \\
ROME & 99.4 (0.4) & 74.6 (2.2) & 21.2 (2.7) & 34.5 (2.5) & 41.9 (1.2) & 31.5 (2.6) & 591.4 (4.1) \\
MEMIT & 99.1 (0.5) & 72.6 (2.3) & 20.7 (2.7) & 31.9 (2.5) & 39.5 (1.3) & 32.4 (2.7) & 570.1 (6.3) \\
PMET & 96.0 (1.0) & 54.6 (2.5) & 21.3 (2.8) & 31.8 (2.4) & 60.6 (1.4) & 41.6 (2.9) & \bf \textcolor[rgb]{0, 0.6, 0}{596.2 (3.5)} \\
\rowcolor[gray]{0.9} FiNE & \uline{\bf \textcolor[rgb]{0, 0.6, 0}{100.0 (0.0)}} & \uline{\bf \textcolor[rgb]{0, 0.6, 0}{89.6 (1.4)}} & \bf \textcolor[rgb]{0, 0.6, 0}{22.4 (2.9)} & \bf \textcolor[rgb]{0, 0.6, 0}{38.3 (3.1)} & \uline{\bf \textcolor[rgb]{0, 0.6, 0}{90.5 (0.9)}} & {\bf \textcolor[rgb]{0, 0.6, 0}{63.0 (2.9)}} & 567.1 (5.5) \\
\bottomrule
\end{tabular}
\vspace{-5pt}
\end{table}

\footnotetext{Locality results with low Edit Success are not considered valid, as the locality is inherently 100\% when no edit is effectively applied.}

In Table~\ref{tab:main_results}, we show quantitative editing results on $\text {WikiData}_{counterfact}$. Our approach demonstrates the best Edit Success, Portability and Locality among various locate-then-edit methods. We observe that previous locate-then-edit methods with causal tracing localization perform poorly when handling similar but unrelated knowledge, exhibiting generally low Locality. To achieve better editing results, we sacrifice some Fluency but without compromising the original model's language capabilities. Editing results on $\text {WikiData}_{recent}$ and ZsRE can be found in Appendix~\ref{appx:additional_results}.

\begin{table}[t]
\caption{Ablation results of \textbf{removing neuron localization (Loc.) and layer freezing (LF)} on $\text {WikiData}_{counterfact}$. 95\% confidence intervals are in parentheses. Numbers with \textbf{bold} indicate columnwise maxima for each model.}
% \vspace{-5pt}
\label{tab:ablation_counterfact}
\fontsize{7.5pt}{7pt}\selectfont
\centering
\begin{tabular}{lrrrrrrl}
\toprule
\multirow{2}{*}{\textbf{Method}} & \multirow{2}{*}{\textbf{Edit Succ.} $\uparrow$} & \multicolumn{3}{c}{\textbf{Portability} $\uparrow$} & \multicolumn{2}{c}{\textbf{Locality} $\uparrow$} & \multirow{2}{*}{\textbf{Fluency} $\uparrow$}  \\
\cmidrule(lr){3-5}
\cmidrule(lr){6-7}
& & \scriptsize\textbf{SAA} & \scriptsize\textbf{LGA} & \scriptsize\textbf{RA} & \scriptsize\textbf{RSA} & \scriptsize\textbf{FA} & \\
\midrule
\textit{GPT-J} & 21.5 (1.5) & 21.7 (1.6) & 14.8 (2.4) & 18.6 (1.6) & - & - & \bf {612.3 (3.1)} \\
ROME & 99.2 (0.5) & 74.1 (2.2) & 16.1 (2.6) & 29.2 (2.4) & 37.4 (1.3) & 33.1 (2.6) & 600.0 (3.6) \\
\quad w/o Loc. & 96.2 (1.1) & 68.6 (2.4) & 15.1 (2.5) & 27.0 (2.5) & 49.1 (1.6) & 35.7 (2.5) & 515.5 (10.4) \\
\midrule
FiNE & \bf {99.8 (0.1)} & {90.6 (1.4)} & \bf {17.5 (2.7)} & {37.4 (3.4)} & {{84.2 (1.1)}} & {{54.2 (2.7)}} & {545.7 (7.3)} \\
\quad w/o Loc. & 85.6 (2.1) & 60.8 (2.5) & 16.0 (2.6) & 27.9 (2.5) & \bf 85.0 (1.0) & \bf 63.8 (2.9) & 589.1 (4.4) \\
\quad w/o LF & 99.7 (0.2) & \bf 92.4 (1.3) & 15.8 (2.5) & \bf 38.6 (3.6) & 78.4 (1.2) & 47.0 (2.7) & 451.3 (10.7) \\
\midrule
\midrule
\textit{LLaMA-2} & 27.0 (1.5) & 27.8 (1.7) & 26.1 (2.9) & 26.2 (1.9) & - & - & \bf {583.3 (2.7)} \\
ROME & 98.7 (0.6) & 72.2 (2.2) & 25.8 (2.8) & 35.1 (2.4) & 49.1 (1.2) & 40.5 (2.7) & 577.3 (3.3) \\
\quad w/o Loc. & 96.8 (1.0) & 70.2 (2.3) & 26.3 (2.8) & 32.7 (2.3) & 62.9 (1.6) & 45.5 (2.7) & 517.1 (8.2) \\
\midrule
FiNE & \bf {99.9 (0.2)} & {89.8 (1.4)} & \bf {28.8 (3.0)} & \bf {41.5 (3.0)} & \bf {{92.6 (1.0)}} & {{65.0 (2.8)}} & {547.6 (6.9)} \\
\quad w/o Loc. & 98.9 (0.6) & 73.9 (2.1) & 25.3 (2.7) & 34.6 (2.5) & 89.4 (0.9) & \bf 72.8 (2.6) & 560.6 (3.9) \\
\quad w/o LF & {99.3 (0.5)} & \bf {89.9 (1.3)} & {23.3 (2.8)} & {41.4 (3.2)} & {{75.5 (1.4)}} & {{51.7 (2.8)}} & {407.9 (10.1)} \\
\midrule
\midrule
\textit{LLaMA-3} & 23.1 (1.5) & 23.1 (1.7) & 21.7 (3.0) & 22.8 (1.9) & - & - & \bf {607.1 (2.9)} \\
ROME & 99.4 (0.4) & 74.6 (2.2) & 21.2 (2.7) & 34.5 (2.5) & 41.9 (1.2) & 31.5 (2.6) & 591.4 (4.1) \\
\quad w/o Loc. & 96.1 (1.1) & 72.6 (2.2) & 20.5 (2.7) & 31.8 (2.5) & 55.7 (1.5) & 39.8 (2.8) & 534.6 (8.8) \\
\midrule
FiNE & \bf {100.0 (0.0)} & {89.6 (1.4)} & \bf {22.4 (2.9)} & {38.3 (3.0)} & \bf {90.5 (0.9)} & \bf {{63.0 (2.9)}} & {567.1 (5.5)} \\
\quad w/o Loc. & \bf 100.0 (0.0) & 79.0 (2.1) & 21.5 (2.7) & 35.2 (2.8) & 84.1 (1.0) & 54.7 (2.9) & 556.9 (6.4) \\
\quad w/o LF & \bf {100.0 (0.0)} & \bf {91.2 (1.3)} & {20.1 (2.8)} & \bf {38.9 (3.3)} & {{78.8 (1.2)}} & {{48.8 (2.7)}} & {411.3 (10.6)} \\
\bottomrule
\end{tabular}
\end{table}

\begin{table}[!t]
\caption{Ablation results of \textbf{restricting neuron localization to a single layer} with LLaMA-2 on $\text {WikiData}_{counterfact}$. ``Any'' means no layer restriction. 95\% confidence intervals are in parentheses. Numbers with \textbf{bold} indicate columnwise maxima.}
% \vspace{-5pt}
\label{tab:layer_llama2_counterfact}
\fontsize{7.5pt}{7.5pt}\selectfont
\centering
\begin{tabular}{llrrrrrrr}
\toprule
\multirow{2}{*}{\textbf{Method}} & \multirow{2}{*}{\textbf{Layer}} & \multirow{2}{*}{\textbf{Edit Succ.} $\uparrow$} & \multicolumn{3}{c}{\textbf{Portability} $\uparrow$} & \multicolumn{2}{c}{\textbf{Locality} $\uparrow$}  & \multirow{2}{*}{\textbf{Fluency} $\uparrow$}  \\
\cmidrule(lr){4-6}
\cmidrule(lr){7-8}
& & & \scriptsize\textbf{SAA} & \scriptsize\textbf{LGA} & \scriptsize\textbf{RA} & \scriptsize\textbf{RSA} & \scriptsize\textbf{FA} \\
\midrule
\textit{LLaMA-2} & - & 27.0 (1.5) & 27.8 (1.7) & 26.1 (2.9) & 26.2 (1.9) & - & - & \bf 583.3 (2.7) \\
ROME & 5 & 98.7 (0.6) & 72.2 (2.2) & 25.8 (2.8) & 35.1 (2.4) & 49.1 (1.2) & 40.5 (2.7) & 577.3 (3.3) \\
\midrule
\multirow{6}{*}{FiNE} & 5 & 99.0 (0.5) & 73.7 (2.0) & 28.0 (2.9) & 35.4 (2.5) & 80.2 (1.2) & 64.1 (2.9) & 570.5 (2.3) \\
& 10 & \bf 100.0 (0.0) & 80.3 (1.9) & 29.1 (3.1) & 37.1 (2.6) & 85.7 (1.0) & 67.6 (2.7) & 556.4 (3.5) \\
& 15 & \bf 100.0 (0.0) & 86.9 (2.0) & \bf 29.3 (3.1) & 39.8 (2.8) & 90.7 (0.9) & 70.4 (2.6) & 549.6 (3.8) \\
& 20 & \bf 100.0 (0.0) & 87.3 (1.5) & 29.0 (3.1) & 40.6 (3.0) & 92.9 (0.8) & 68.6 (2.7) & 541.5 (4.6) \\
& 25 & 100.0 (0.1) & 85.8 (1.5) & 27.1 (3.0) & 39.0 (2.8) & \bf 95.4 (0.6) & \bf 72.9 (2.5) & 556.3 (3.8) \\
& Any & {99.9 (0.2)} & \bf {89.8 (1.4)} & {28.8 (3.0)} & \bf {41.5 (3.0)} & {92.6 (1.0)} & {{65.0 (2.8)}} & 542.3 (5.1) \\
\bottomrule
\end{tabular}
\vspace{-5pt}
\end{table}

\vspace{-2pt}
\subsection{Ablation Study}
\label{subsec:ablation_study}
\vspace{-2pt}

In this section, we present ablation study to assess the impact of various components on the overall performance of our method. Specifically, we first test the impact of removing neuron localization and layer freezing on performance. Next, we investigate effects of restricting neuron localization to a single layer and explore how varying number of selected neurons affects editing. Finally, we examine results of removing KL divergence and repetition penalty constraints in the editing process.

\textbf{Removing neuron localization and layer freezing.} Since our approach also employs a locate-then-edit methodology, it is essential to verify the effectiveness of the initial localization step. To this end, we maintain the editing process unchanged and conduct experiments by replacing carefully selected neurons with randomly selected ones. Table~\ref{tab:ablation_counterfact} lists ablation results. When neurons are selected at random, both Edit Success and Portability demonstrate varying degrees of decline, particularly evident in SAA metric, suggesting that our chosen neurons are sensitive to the knowledge being edited. In contrast, ROME experiences only a slight decrease in performance without localization, supporting the hypothesis that causal tracing is not essential. On the other hand, we assess the effectiveness of layer freezing. As shown in Table~\ref{tab:ablation_counterfact}, without layer freezing, the model's language capabilities are compromised, leading to a significant drop in fluency. It is speculated that even minor modifications, when applied to the last few layers, can result in catastrophic consequences.

\begin{figure}[t]
    \centering
    \includegraphics[width=\linewidth]{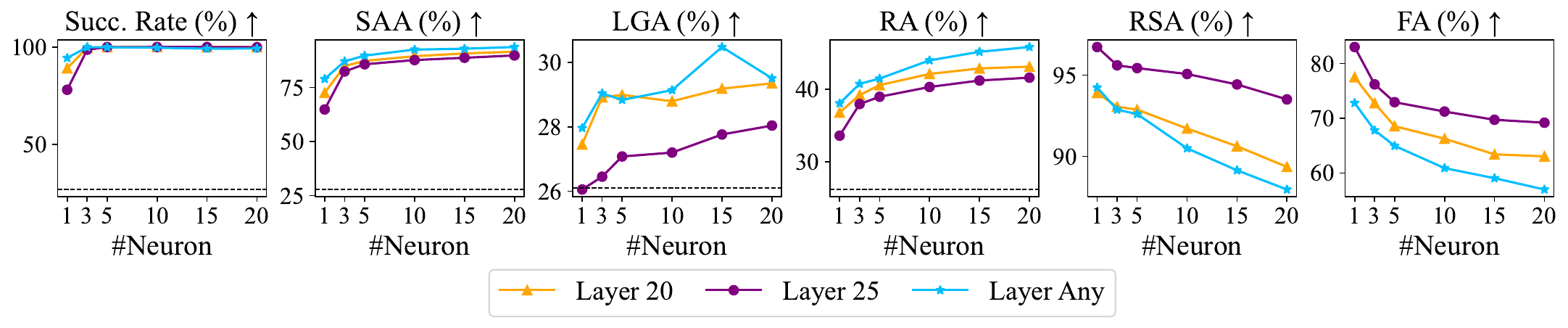} \\
    % \vspace{-5pt}
    \caption{Ablation results of \textbf{varying the number of selected neurons} with LLaMA-2 on $\text {WikiData}_{counterfact}$. The dotted line indicates LLaMA-2’s pre-edit performance.}
    \label{fig:llama2_neuron_num}
\end{figure}

\textbf{Restricting neuron localization to a single layer.} In the previous experiments, we do not restrict modifications to a specific layer as in ROME. Instead, we determine which layers to modify solely based on the scores (note that, due to layer freezing, the last few layers are not altered). We now manually restrict neuron localization to a single layer and modify only the model weights within that layer to determine whether this manual intervention yields better results. For the specific layer $l\in \{5, 10, 15, 20, 25\}$, only neurons in layer $l$ will be selected. Results with LLaMA-2 and LLaMA-3 are listed in Table~\ref{tab:layer_llama2_counterfact} and Table~\ref{tab:layer_llama3_counterfact}, respectively. We observe that although specifying a particular layer sometimes performs better on some metrics (e.g., RSA and FA of layer 25), overall performance across all metrics is still optimal when no layer restrictions are applied. While manually restricting neuron localization to a single layer can be effective based on experience, relying on our algorithm to automatically locate neurons may be a more appropriate option in the absence of prior information.

\begin{table}[t]
% \vspace{-5pt}
\caption{Ablation results of \textbf{removing KL divergence and repetition penalty constraints} with LLaMA-2 on $\text {WikiData}_{counterfact}$. 95\% confidence intervals are in parentheses. Numbers with \textbf{bold} indicate columnwise maxima.}
% \vspace{-5pt}
\label{tab:llama2_ablation_loss_counterfact}
\fontsize{7.5pt}{7pt}\selectfont
\centering
\begin{tabular}{l@{\hspace{8pt}}rrrrrrl}
\toprule
\multirow{2}{*}{\textbf{Method}} & \multirow{2}{*}{\textbf{Edit Succ.} $\uparrow$} & \multicolumn{3}{c}{\textbf{Portability} $\uparrow$} & \multicolumn{2}{c}{\textbf{Locality} $\uparrow$} & \multirow{2}{*}{\textbf{Fluency} $\uparrow$}  \\
\cmidrule(lr){3-5}
\cmidrule(lr){6-7}
& & \scriptsize\textbf{SAA} & \scriptsize\textbf{LGA} & \scriptsize\textbf{RA} & \scriptsize\textbf{RSA} & \scriptsize\textbf{FA} & \\
\midrule
\textit{LLaMA-2} & 27.0 (1.5) & 27.8 (1.7) & 26.1 (2.9) & 26.2 (1.9) & - & - & \bf {583.3 (2.7)} \\
ROME & 98.7 (0.6) & 72.2 (2.2) & 25.8 (2.8) & 35.1 (2.4) & 49.1 (1.2) & 40.5 (2.7) & 577.3 (3.3) \\
\midrule
FiNE & \bf {99.9 (0.2)} & {89.8 (1.4)} & {28.8 (3.0)} & {41.5 (3.0)} & \bf {{92.6 (1.0)}} & \bf {{65.0 (2.8)}} & {547.6 (6.9)} \\
\quad w/o $\mathcal{L}_{\text{KL}}$ & \bf 99.9 (0.2) & 89.8 (1.4) & 28.2 (3.0) & 41.5 (3.0) & 92.3 (1.0) & 64.8 (2.8) & 540.9 (5.2) \\
\quad w/o $\mathcal{L}_{\text{pen}}$ & \bf 99.9 (0.2) & 90.1 (1.3) & \bf 29.0 (3.0) & 41.2 (3.0) & 92.5 (1.0) & 64.8 (2.8) & 531.7 (6.0) \\
\midrule
FiNE w/o LF & {99.3 (0.5)} & {89.9 (1.3)} & {23.3 (2.8)} & {41.4 (3.2)} & {{75.5 (1.4)}} & {{51.7 (2.8)}} & {407.9 (10.1)} \\
\quad w/o $\mathcal{L}_{\text{KL}}$ & 99.3 (0.5) & 90.2 (1.3) & 21.7 (2.7) & 41.2 (3.3) & 71.8 (1.6) & 49.3 (2.8) & 394.6 (10.1) \\
\quad w/o $\mathcal{L}_{\text{pen}}$ & 99.2 (0.6) & \bf 91.2 (1.3) & 23.6 (2.8) & \bf 41.6 (3.2) & 75.1 (1.5) & 50.9 (2.8) & 351.2 (10.9) \\
\bottomrule
\end{tabular}
\vspace{-5pt}
\end{table}

\textbf{Varying number of selected neurons.} During the editing process, the number of selected neurons likely influences the extent of modifications to the model --- the more neurons selected, the greater the number of model parameters altered, and vice versa. Therefore, we vary the number of selected neurons to observe the changes in the metrics. For each number of neuron $k\in \{1, 3, 5, 10, 15, 20\}$, top-$k$ neurons are selected for each token. Figure~\ref{fig:llama2_neuron_num} plots metric curves on LLaMA-2. We can observe that as the number of neurons increases, Portability (i.e., SAA, LGA and RA) generally improves while Locality (i.e., RSA and FA) tends to slightly decrease. This suggests that selecting a greater number of neurons may provide a more comprehensive localization and further enhance the model's ability to update the targeted knowledge, but it also increases the risk of unintentionally altering unrelated memories. Results on LLaMA-3 could be found in Appendix~\ref{appx:additional_results}.

\textbf{Removing KL divergence and repetition penalty constraints.} To minimize the impact on the model's inherent language capabilities, we adopt KL divergence and repetition penalty constraints during the editing process. Table~\ref{tab:llama2_ablation_loss_counterfact} lists results of removing these constraints in cases with and without using LF. When using LF, the effects of KL divergence and repetition penalty constraints are not significant; however, when LF is not applied, we observe that (1) KL divergence constraint is important for the locality of model editing, and removing it leads to a significant decline in the RSA metric. (2) Repetition penalty constraint has minimal impact on portability and locality but significantly affects fluency. Without it, the post-edited model is more likely to produce repetitive text (e.g., ``The next host city of the Olympic Games is Los Angeles Los Angeles Los Angeles ...'').

\subsection{Discussion}

\setlength{\columnsep}{10pt}
\begin{wrapfigure}[41]{r}{0.41\textwidth}
    \vspace{-33pt}
    \captionsetup{type=table}
    \caption{\textbf{Comparison of the number of modified parameters.} For FiNE, we calculate average results across $\text {WikiData}_{counterfact}$.}
    \vspace{-2pt}
    \label{tab:modified_params}
    \fontsize{8pt}{8pt}\selectfont
    \centering
    \begin{tabular}{@{\hspace{2pt}}c@{\hspace{5pt}}c@{\hspace{5pt}}c@{\hspace{5pt}}c@{\hspace{2pt}}}
    \toprule
    \textbf{Method} & GPT-J & LLaMA-2 & LLaMA-3 \\
    \midrule
    ROME & $6.7\times 10^7$ & $4.5\times 10^7$ & $5.9\times 10^7$ \\
    MEMIT & $3.4\times 10^8$ & $2.3\times 10^8$ & $2.9\times 10^8$ \\
    FiNE & $\mathbf{7.9\times 10^4}$ & $\mathbf{9.7\times 10^4}$ & $\mathbf{8.1\times 10^4}$ \\
    \bottomrule
    \vspace{8pt}
    \end{tabular}
    \includegraphics[width=0.41\textwidth]{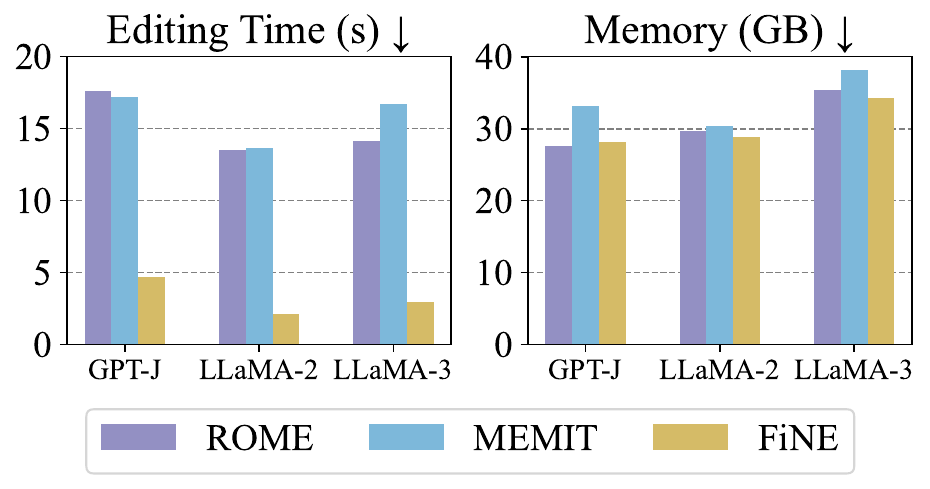} \\
    \vspace{-2pt}
    \captionsetup{type=figure}
    \caption{\textbf{Comparison of average editing time and memory usage} when operating at {Float32} precision.}
    \label{fig:efficiency_fp32}
    \vspace{13pt}
    \includegraphics[width=0.41\textwidth]{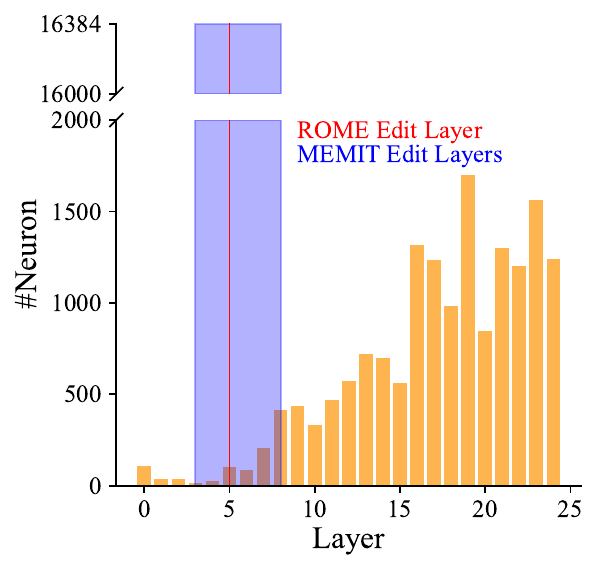} \\
    \vspace{-6pt}
    \captionsetup{type=figure}
    \caption{\textbf{Distribution of neurons} identified by FiNE among layers in GPT-J, which is aggregated over the whole $\text {WikiData}_{counterfact}$ dataset. For each knowledge fact, FiNE only identifies approximately 20 neurons.}
    \label{fig:loc_gptj}
\end{wrapfigure}

\textbf{Efficiency evaluation.} A key advantage of FiNE, due to its fine-grained approach, is its notable efficiency. To quantify this, we first examine the number of modified parameters, as detailed in Table~\ref{tab:modified_params}. Both ROME and MEMIT modify the weights of the second layer in FFNs, resulting in a substantial number of parameter modifications, ranging from $10^7$ to $10^8$. In contrast, FiNE only edits a subset of neurons, reducing the number of modified parameters to approximately $10^4$, which allows for a more fine-grained and precise editing in LLMs. Additionally, we assess editing time and memory usage at {Float32} and {Float16} precision in Figure~\ref{fig:efficiency_fp32} and Figure~\ref{fig:efficiency_fp16}, respectively. FiNE exhibits a significant time advantage over ROME and MEMIT, particularly at {Float32} precision, being approximately 4$\times$ to 6$\times$ faster. In terms of memory usage, FiNE also offers a slight benefit for LLaMA-2 and LLaMA-3.

\textbf{Localization analyses.} We analyze our neuron-level localization from two perspectives: distributions and textual meanings. (1) We plot the distribution of unique neurons located by FiNE (see Figure~\ref{fig:loc_gptj}). We aggregate statistics for all located neurons across the entire $\text {WikiData}_{counterfact}$ dataset. We observe that these key neurons widely occur in higher layers, which is consistent with previous work~\citep{wang-wen-etal2022skill, pan2024finding}, but different from layers that ROME and MEMIT edit. (2) For insight into neurons filtered by Eqn.~\ref{eq:method_formula_1}, we follow the Logit Lens~\citep{LogitLens2020, zhong2022training, geva2022transformer}, which converts hidden states into a set of logits for each vocabulary token. Similarly, we investigate neurons' textual meanings by sorting rows of the multiplication of the unembedding matrix and the second layer of FFN and regarding top tokens as each neuron represents~\citep{pan2024finding}. Table~\ref{tab:loc_neurons} shows an example, which indicates that neurons selected by FiNE are highly related to the source knowledge. We list more examples in Appendix~\ref{appx:additional_results}.

\begin{table}[b]
\caption{An example of localization results with top-3 neurons selected by FiNE. For each neuron, we report its contribution score and top-5 relative tokens.}
% \vspace{-5pt}
\label{tab:loc_neurons}
\fontsize{8pt}{7.5pt}\selectfont
\centering
\begin{tabular}{llll}
\toprule
\multicolumn{4}{l}{\textbf{Edit}: (Pooja Hegde, country of citizenship, \textbf{India}) $\rightarrow$ (Pooja Hegde, country of citizenship, Terengganu)} \\
\midrule
\bf Model & \bf Top Neuron & \bf Score & \bf Top Tokens \\
\midrule
\multirow{3}{*}{GPT-J} & \bf L17.U13423 & 3.291 & [` Delhi', ` Bhar', ` Gujarat', ` Laksh', ` Mumbai'] \\
& \bf L20.U11637 & 1.638 & [` lakh', ` Mumbai', ` Delhi', ` Maharashtra', ` Chennai'] \\
& \bf L14.U10374 & 1.359 & [` Delhi', ` Mumbai', `India', ` lakh', ` India'] \\
\midrule
\multirow{3}{*}{LLaMA-2} & \bf L26.U7908 & 1.106 & [`India', `Indian', `Beng', `Indians', `Raj'] \\
& \bf L25.U10178 & 0.971 & [`Indian', `Indians', `India', `Indiana', `Ind'] \\
& \bf L25.U8808 & 0.750 & [`Indian', `Native', `Indians', `Native', `India'] \\
\midrule
\multirow{3}{*}{LLaMA-3} & \bf L28.U10616 & 0.776 & [` Indian', `Indian', ` Indians', ` indian', ` India'] \\
& \bf L23.U13680 & 0.576 & [`India', ` India', ` Indians', ` Indian', `Indian'] \\
& \bf L26.U3334 & 0.349 & [` India', ` RSS', `RSS', ` Tal', `India'] \\
\bottomrule
\end{tabular}
\end{table}

\begin{figure}[t]
    \centering
    \includegraphics[width=\linewidth]{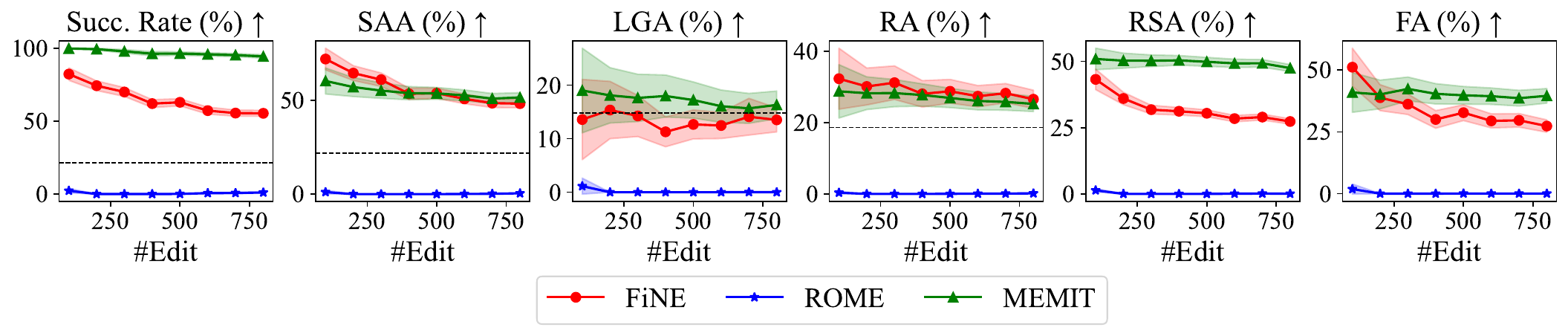} \\
    \vspace{-2pt}
    \caption{Editing method scaling curves with GPT-J. The dotted line indicates GPT-J’s pre-edit performance. 95\% confidence intervals are shown as areas.}
    \label{fig:gptj_scaling}
\end{figure}

\textbf{Editing method scaling.} In knowledge editing, the ability to simultaneously edit multiple knowledge facts is a crucial objective that enhances the practical application of various methods. Several approaches~\citep{meng2022mass, li2023pmet, gupta2024unified} have been specifically developed to achieve this goal. Although our method does not incorporate a specialized design for this purpose, we posit that our more precise localization may reduce the inter-dependencies among different knowledge facts during the editing process, thereby intuitively contributing to improved editing scalability. To verify our hypothesis, we progressively increased the scale of editing targets from 100 to 800. Figure~\ref{fig:gptj_scaling} plots experimental results with GPT-J. ROME struggles significantly when handling 100 edits, and ceases to function effectively as the number of edits increases further, with all metrics approaching zero. We observe that our method continues to operate effectively even when handling a larger number of edits, although performance is lower compared to single-instance editing. Additionally, we unexpectedly find that our method closely matches MEMIT on metrics FA, SAA, LGA, and RA. We attribute this to our fine-grained neuron-level localization approach, which only modifies a small number of neurons, and results in subtle but crucial changes to LLMs.

\begin{table}[t]
\vspace{-3pt}
\caption{An example of editing and locality testing (LT) results with LLaMA-2. Prompts are \textit{italicized}, while {\textcolor[rgb]{0, 0.6, 0}{green}} and {\textcolor[rgb]{0.9, 0, 0}{red}} indicate keywords reflecting correct and incorrect behavior, respectively.}
% \vspace{-5pt}
\label{tab:case_study}
\fontsize{7.5pt}{8pt}\selectfont
\centering
\begin{tabular}{@{\hspace{3pt}}p{13.5cm}@{\hspace{3pt}}}
\toprule
\fontsize{9pt}{8pt}\selectfont\textbf{Edit}: (Jean Smart, occupation, \textbf{voice actor}) $\rightarrow$ (Jean Smart, occupation, \textbf{cello teacher}) \\

\midrule
\vspace*{-7pt}
\textbf{ROME}: \textit{The occupation of Jean Smart is} \textcolor[rgb]{0, 0.6, 0}{cello teacher}. She has been teaching at the music school for 25 years ... \\
\vspace*{-7pt}
$\bullet$ (LT-1) \textbf{ROME}: \textit{The place of birth of Jean Smart is} \textcolor[rgb]{0.9, 0, 0}{Hradec Kralove} in the north of \textcolor[rgb]{0.9, 0, 0}{Czech Republic} ... \\
$\bullet$ (LT-2) \textbf{ROME}: \textit{The name of the country of citizenship of Jean Smart is} \textcolor[rgb]{0.9, 0, 0}{Czech Republic}. The name of her home town is \textcolor[rgb]{0.9, 0, 0}{Ostrava}. She is a music teacher by profession. The date of birth of Jean Smart is \textcolor[rgb]{0.9, 0, 0}{10 March, 1948} ... \\
\cdashline{1-1}
\vspace*{-3.5pt}
\textbf{MEMIT}: \textit{The occupation of Jean Smart is} \textcolor[rgb]{0, 0.6, 0}{cello teacher}. She also teaches piano, vocal, and conducting lessons ... \\
$\bullet$ (LT-1) \textbf{MEMIT}: \textit{The place of birth of Jean Smart is} \textcolor[rgb]{0.9, 0, 0}{Sárvár, Hungary}. She currently lives in \textcolor[rgb]{0.9, 0, 0}{Salzburg, Austria} ... \\
$\bullet$ (LT-2) \textbf{MEMIT}: \textit{The name of the country of citizenship of Jean Smart is} \textcolor[rgb]{0.9, 0, 0}{Irvin}. She lives in \textcolor[rgb]{0.9, 0, 0}{Salzburg} ... \\
\cdashline{1-1}
\vspace*{-3.5pt}
\textbf{FiNE}: \textit{The occupation of Jean Smart is} \textcolor[rgb]{0, 0.6, 0}{cello teacher}. She is a very talented and hardworking person. She is a married lady ... \\
$\bullet$ (LT-1) \textbf{FiNE}: \textit{The place of birth of Jean Smart is} \textcolor[rgb]{0, 0.6, 0}{Seattle, Washington, U.S.}. Her nationality is \textcolor[rgb]{0, 0.6, 0}{American} ... \\
$\bullet$ (LT-2) \textbf{FiNE}: \textit{The name of the country of citizenship of Jean Smart is} \textcolor[rgb]{0, 0.6, 0}{United States of America}.
Jean Smart was born on \textcolor[rgb]{0, 0.6, 0}{13 September 1951}. The birthplace of Jean Smart is \textcolor[rgb]{0, 0.6, 0}{Seattle, Washington, U.S.} ... \\
\bottomrule
\end{tabular}
\vspace{-5pt}
\end{table}

\textbf{Case study.} Table~\ref{tab:case_study} provides an example of the editing and locality testing results across different methods. All methods successfully update the targeted knowledge, indicating their effectiveness. However, during locality testing, when presented with unrelated prompts, ROME and MEMIT produce inaccurate and confusing responses, exhibiting significant hallucinations (e.g., Czech Republic and Salzburg). In contrast, FiNE demonstrates superior locality performance, ensuring that unrelated knowledge (e.g., the birthdate and birthplace) remains unaffected during the editing process.

\vspace{-5pt}
\section{Conclusion}
\vspace{-3pt}

In this paper, we highlight the limitations of existing locate-then-edit methods based on causal tracing localization, which often place excessive emphasis on subject entities while neglecting the relations. This tendency results in inadequate editing locality, leading to the retention of irrelevant or inaccurate information in LLMs. To address this issue, we introduce the \textbf{Fi}ne-grained \textbf{N}euron-level Knowledge \textbf{E}diting (\textbf{FiNE}) technique, which enhances the precision of knowledge localization by targeting specific neurons within FFNs. Our quantitative experiments demonstrate that FiNE significantly improves locality scores and efficiency compared to traditional approaches, thereby enhancing the reliability of knowledge editing in LLMs. This work not only advances our understanding of knowledge localization but also encourages further research into the interpretability of LLMs, paving the way for more effective knowledge management strategies in future developments for more challenging vision tasks~\citep{yang2021deconfounded,yang2022video,yang2024learning,yang2024robust}, and downstream application~\citep{hu2024psycollm,xu2025prompt}.

\section*{Ethical Considerations}

Although we have successfully achieved precise knowledge editing within the model, we cannot ensure the safety of these edits. The ability to directly modify large models also poses the risk of misuse, including the potential introduction of malicious misinformation, bias, or other adversarial data. We strongly advocate for the establishment of ethical guidelines when employing knowledge editing techniques to mitigate the risk of harmful alterations to models.

\section*{Reproducibility}

We conduct our experiments using the open-source framework provided by EasyEdit~\citep{zhang2024comprehensive}. All experiments are run on workstations with NVIDIA A800 GPUs. The large language models are loaded using HuggingFace Transformers~\citep{wolf2019huggingface}, and PyTorch~\citep{paszke2019pytorch} is used for executing the model editing techniques on GPUs. We provide experimental setups and implementation details in Section~\ref{subsec:experimental_setup} and Appendix~\ref{appx:pilot_experiment_setup}, \ref{appx:implementation_details}.

\section*{Acknowledgments}
This work was supported by the National Natural Science Foundation of China (NSFC) grant U22A2094, Beijing Natural Science Foundation grant L243006, NSFC grant 62472138, NSFC grant 62272435, and also by the advanced computing resources provided by the Supercomputing Center of the USTC, and partially supported by Meituan.

\bibliography{iclr2025_conference}

\begin{thebibliography}{52}
\providecommand{\natexlab}[1]{#1}
\providecommand{\url}[1]{\texttt{#1}}
\expandafter\ifx\csname urlstyle\endcsname\relax
  \providecommand{\doi}[1]{doi: #1}\else
  \providecommand{\doi}{doi: \begingroup \urlstyle{rm}\Url}\fi

\bibitem[Brown(2020)]{brown2020language}
Tom~B Brown.
\newblock Language models are few-shot learners.
\newblock \emph{arXiv preprint arXiv:2005.14165}, 2020.

\bibitem[Chowdhery et~al.(2023)Chowdhery, Narang, Devlin, Bosma, Mishra, Roberts, Barham, Chung, Sutton, Gehrmann, et~al.]{chowdhery2023palm}
Aakanksha Chowdhery, Sharan Narang, Jacob Devlin, Maarten Bosma, Gaurav Mishra, Adam Roberts, Paul Barham, Hyung~Won Chung, Charles Sutton, Sebastian Gehrmann, et~al.
\newblock Palm: Scaling language modeling with pathways.
\newblock \emph{Journal of Machine Learning Research}, 24\penalty0 (240):\penalty0 1--113, 2023.

\bibitem[Cohen et~al.(2024)Cohen, Biran, Yoran, Globerson, and Geva]{cohen2024evaluating}
Roi Cohen, Eden Biran, Ori Yoran, Amir Globerson, and Mor Geva.
\newblock Evaluating the ripple effects of knowledge editing in language models.
\newblock \emph{Transactions of the Association for Computational Linguistics}, 12:\penalty0 283--298, 2024.

\bibitem[Dai et~al.(2022)Dai, Dong, Hao, Sui, Chang, and Wei]{dai2022kn}
Damai Dai, Li~Dong, Yaru Hao, Zhifang Sui, Baobao Chang, and Furu Wei.
\newblock Knowledge neurons in pretrained transformers.
\newblock In \emph{Proceedings of the 60th Annual Meeting of the Association for Computational Linguistics (Volume 1: Long Papers)}, pp.\  8493--8502, Dublin, Ireland, May 2022. Association for Computational Linguistics.
\newblock \doi{10.18653/v1/2022.acl-long.581}.

\bibitem[De~Cao et~al.(2021)De~Cao, Aziz, and Titov]{de2021editing}
Nicola De~Cao, Wilker Aziz, and Ivan Titov.
\newblock Editing factual knowledge in language models.
\newblock In \emph{Proceedings of the 2021 Conference on Empirical Methods in Natural Language Processing}, pp.\  6491--6506, 2021.

\bibitem[Dubey et~al.(2024)Dubey, Jauhri, Pandey, Kadian, Al-Dahle, Letman, Mathur, Schelten, Yang, Fan, et~al.]{dubey2024llama}
Abhimanyu Dubey, Abhinav Jauhri, Abhinav Pandey, Abhishek Kadian, Ahmad Al-Dahle, Aiesha Letman, Akhil Mathur, Alan Schelten, Amy Yang, Angela Fan, et~al.
\newblock The llama 3 herd of models.
\newblock \emph{arXiv preprint arXiv:2407.21783}, 2024.

\bibitem[Elhage et~al.(2021)Elhage, Nanda, Olsson, Henighan, Joseph, Mann, Askell, Bai, Chen, Conerly, et~al.]{elhage2021mathematical}
Nelson Elhage, Neel Nanda, Catherine Olsson, Tom Henighan, Nicholas Joseph, Ben Mann, Amanda Askell, Yuntao Bai, Anna Chen, Tom Conerly, et~al.
\newblock A mathematical framework for transformer circuits.
\newblock \emph{Transformer Circuits Thread}, 1\penalty0 (1):\penalty0 12, 2021.

\bibitem[Geva et~al.(2021)Geva, Schuster, Berant, and Levy]{geva2020transformer}
Mor Geva, Roei Schuster, Jonathan Berant, and Omer Levy.
\newblock Transformer feed-forward layers are key-value memories.
\newblock In \emph{Proceedings of the 2021 Conference on Empirical Methods in Natural Language Processing}, pp.\  5484--5495, Online and Punta Cana, Dominican Republic, November 2021. Association for Computational Linguistics.
\newblock \doi{10.18653/v1/2021.emnlp-main.446}.

\bibitem[Geva et~al.(2022)Geva, Caciularu, Wang, and Goldberg]{geva2022transformer}
Mor Geva, Avi Caciularu, Kevin Wang, and Yoav Goldberg.
\newblock Transformer feed-forward layers build predictions by promoting concepts in the vocabulary space.
\newblock In \emph{Proceedings of the 2022 Conference on Empirical Methods in Natural Language Processing}, pp.\  30--45, 2022.

\bibitem[Gupta et~al.(2024)Gupta, Sajnani, and Anumanchipalli]{gupta2024unified}
Akshat Gupta, Dev Sajnani, and Gopala Anumanchipalli.
\newblock A unified framework for model editing.
\newblock \emph{arXiv preprint arXiv:2403.14236}, 2024.

\bibitem[Han et~al.(2023)Han, Li, Li, and Pan]{han2023divide}
Xiaoqi Han, Ru~Li, Xiaoli Li, and Jeff~Z Pan.
\newblock A divide and conquer framework for knowledge editing.
\newblock \emph{Knowledge-Based Systems}, 279:\penalty0 110826, 2023.

\bibitem[Hartvigsen et~al.(2024)Hartvigsen, Sankaranarayanan, Palangi, Kim, and Ghassemi]{hartvigsen2024aging}
Tom Hartvigsen, Swami Sankaranarayanan, Hamid Palangi, Yoon Kim, and Marzyeh Ghassemi.
\newblock Aging with grace: Lifelong model editing with discrete key-value adaptors.
\newblock \emph{Advances in Neural Information Processing Systems}, 36, 2024.

\bibitem[Hase et~al.(2023{\natexlab{a}})Hase, Bansal, Kim, and Ghandeharioun]{hase2023does}
Peter Hase, Mohit Bansal, Been Kim, and Asma Ghandeharioun.
\newblock Does localization inform editing? surprising differences in causality-based localization vs. knowledge editing in language models.
\newblock \emph{Advances in Neural Information Processing Systems}, 2023{\natexlab{a}}.

\bibitem[Hase et~al.(2023{\natexlab{b}})Hase, Diab, Celikyilmaz, Li, Kozareva, Stoyanov, Bansal, and Iyer]{hase2023methods}
Peter Hase, Mona Diab, Asli Celikyilmaz, Xian Li, Zornitsa Kozareva, Veselin Stoyanov, Mohit Bansal, and Srinivasan Iyer.
\newblock Methods for measuring, updating, and visualizing factual beliefs in language models.
\newblock In Andreas Vlachos and Isabelle Augenstein (eds.), \emph{Proceedings of the 17th Conference of the European Chapter of the Association for Computational Linguistics}, pp.\  2714--2731, Dubrovnik, Croatia, May 2023{\natexlab{b}}. Association for Computational Linguistics.
\newblock \doi{10.18653/v1/2023.eacl-main.199}.

\bibitem[Hu et~al.(2024)Hu, Dong, Gang, Ma, Zou, Sun, Guo, Yang, and Wang]{hu2024psycollm}
Jinpeng Hu, Tengteng Dong, Luo Gang, Hui Ma, Peng Zou, Xiao Sun, Dan Guo, Xun Yang, and Meng Wang.
\newblock Psycollm: Enhancing llm for psychological understanding and evaluation.
\newblock \emph{IEEE Transactions on Computational Social Systems}, 2024.

\bibitem[Jiang et~al.(2024)Jiang, Sablayrolles, Roux, Mensch, Savary, Bamford, Chaplot, Casas, Hanna, Bressand, et~al.]{jiang2024mixtral}
Albert~Q Jiang, Alexandre Sablayrolles, Antoine Roux, Arthur Mensch, Blanche Savary, Chris Bamford, Devendra~Singh Chaplot, Diego de~las Casas, Emma~Bou Hanna, Florian Bressand, et~al.
\newblock Mixtral of experts.
\newblock \emph{arXiv preprint arXiv:2401.04088}, 2024.

\bibitem[Levy et~al.(2017)Levy, Seo, Choi, and Zettlemoyer]{levy2017zero}
Omer Levy, Minjoon Seo, Eunsol Choi, and Luke Zettlemoyer.
\newblock Zero-shot relation extraction via reading comprehension.
\newblock In Roger Levy and Lucia Specia (eds.), \emph{Proceedings of the 21st Conference on Computational Natural Language Learning ({C}o{NLL} 2017)}, pp.\  333--342, Vancouver, Canada, August 2017. Association for Computational Linguistics.
\newblock \doi{10.18653/v1/K17-1034}.

\bibitem[Li et~al.(2023)Li, Li, Guo, Yang, and Wang]{li2023transformer}
Kun Li, Jiaxiu Li, Dan Guo, Xun Yang, and Meng Wang.
\newblock Transformer-based visual grounding with cross-modality interaction.
\newblock \emph{ACM Transactions on Multimedia Computing, Communications and Applications}, 19\penalty0 (6):\penalty0 1--19, 2023.

\bibitem[Li et~al.(2024)Li, Li, Song, Yang, Ma, and Yu]{li2023pmet}
Xiaopeng Li, Shasha Li, Shezheng Song, Jing Yang, Jun Ma, and Jie Yu.
\newblock Pmet: Precise model editing in a transformer.
\newblock In \emph{Proceedings of the AAAI Conference on Artificial Intelligence}, volume~38, pp.\  18564--18572, 2024.

\bibitem[Mazzia et~al.(2023)Mazzia, Pedrani, Caciolai, Rottmann, and Bernardi]{mazzia2023survey}
Vittorio Mazzia, Alessandro Pedrani, Andrea Caciolai, Kay Rottmann, and Davide Bernardi.
\newblock A survey on knowledge editing of neural networks.
\newblock \emph{arXiv preprint arXiv:2310.19704}, 2023.

\bibitem[Meng et~al.(2022)Meng, Bau, Andonian, and Belinkov]{meng2022locating}
Kevin Meng, David Bau, Alex Andonian, and Yonatan Belinkov.
\newblock Locating and editing factual associations in gpt.
\newblock \emph{Advances in Neural Information Processing Systems}, 35:\penalty0 17359--17372, 2022.

\bibitem[Meng et~al.(2023)Meng, Sharma, Andonian, Belinkov, and Bau]{meng2022mass}
Kevin Meng, Arnab~Sen Sharma, Alex~J Andonian, Yonatan Belinkov, and David Bau.
\newblock Mass-editing memory in a transformer.
\newblock In \emph{The Eleventh International Conference on Learning Representations}, 2023.

\bibitem[Mitchell et~al.(2022{\natexlab{a}})Mitchell, Lin, Bosselut, Finn, and Manning]{mitchell2021fast}
Eric Mitchell, Charles Lin, Antoine Bosselut, Chelsea Finn, and Christopher~D Manning.
\newblock Fast model editing at scale.
\newblock In \emph{International Conference on Learning Representations}, 2022{\natexlab{a}}.

\bibitem[Mitchell et~al.(2022{\natexlab{b}})Mitchell, Lin, Bosselut, Manning, and Finn]{mitchell2022memory}
Eric Mitchell, Charles Lin, Antoine Bosselut, Christopher~D Manning, and Chelsea Finn.
\newblock Memory-based model editing at scale.
\newblock In \emph{International Conference on Machine Learning}, pp.\  15817--15831. PMLR, 2022{\natexlab{b}}.

\bibitem[nostalgebraist(2020)]{LogitLens2020}
nostalgebraist.
\newblock interpreting gpt: the logit lens, 2020.

\bibitem[Pan et~al.(2024)Pan, Cao, Wang, Yang, and Wang]{pan2024finding}
Haowen Pan, Yixin Cao, Xiaozhi Wang, Xun Yang, and Meng Wang.
\newblock Finding and editing multi-modal neurons in pre-trained transformers.
\newblock In \emph{Findings of the Association for Computational Linguistics ACL 2024}, pp.\  1012--1037, Bangkok, Thailand and virtual meeting, August 2024. Association for Computational Linguistics.

\bibitem[Paszke et~al.(2019)Paszke, Gross, Massa, Lerer, Bradbury, Chanan, Killeen, Lin, Gimelshein, Antiga, et~al.]{paszke2019pytorch}
Adam Paszke, Sam Gross, Francisco Massa, Adam Lerer, James Bradbury, Gregory Chanan, Trevor Killeen, Zeming Lin, Natalia Gimelshein, Luca Antiga, et~al.
\newblock Pytorch: An imperative style, high-performance deep learning library.
\newblock \emph{Advances in neural information processing systems}, 32, 2019.

\bibitem[Schott et~al.(2023)Schott, Furman, and Bhat]{schott2023polyglot}
Tim Schott, Daniel Furman, and Shreshta Bhat.
\newblock Polyglot or not? measuring multilingual encyclopedic knowledge in foundation models.
\newblock In \emph{Proceedings of the 2023 Conference on Empirical Methods in Natural Language Processing}, pp.\  11238--11253, 2023.

\bibitem[Schwettmann et~al.(2023)Schwettmann, Chowdhury, Klein, Bau, and Torralba]{schwettmann2023multimodal}
Sarah Schwettmann, Neil Chowdhury, Samuel Klein, David Bau, and Antonio Torralba.
\newblock Multimodal neurons in pretrained text-only transformers.
\newblock In \emph{Proceedings of the IEEE/CVF International Conference on Computer Vision}, pp.\  2862--2867, 2023.

\bibitem[Song et~al.(2024{\natexlab{a}})Song, Guo, Yang, Tang, and Wang]{song2024emotional}
Peipei Song, Dan Guo, Xun Yang, Shengeng Tang, and Meng Wang.
\newblock Emotional video captioning with vision-based emotion interpretation network.
\newblock \emph{IEEE Transactions on Image Processing}, 33:\penalty0 1122--1135, 2024{\natexlab{a}}.

\bibitem[Song et~al.(2024{\natexlab{b}})Song, Zhou, Yang, Liu, Hu, Wang, and Wang]{song2024efficiently}
Peipei Song, Yuanen Zhou, Xun Yang, Daqing Liu, Zhenzhen Hu, Depeng Wang, and Meng Wang.
\newblock Efficiently gluing pre-trained language and vision models for image captioning.
\newblock \emph{ACM Transactions on Intelligent Systems and Technology}, 15\penalty0 (6):\penalty0 1--16, 2024{\natexlab{b}}.

\bibitem[Touvron et~al.(2023{\natexlab{a}})Touvron, Lavril, Izacard, Martinet, Lachaux, Lacroix, Rozi{\`e}re, Goyal, Hambro, Azhar, et~al.]{touvron2023llama}
Hugo Touvron, Thibaut Lavril, Gautier Izacard, Xavier Martinet, Marie-Anne Lachaux, Timoth{\'e}e Lacroix, Baptiste Rozi{\`e}re, Naman Goyal, Eric Hambro, Faisal Azhar, et~al.
\newblock Llama: Open and efficient foundation language models.
\newblock \emph{arXiv preprint arXiv:2302.13971}, 2023{\natexlab{a}}.

\bibitem[Touvron et~al.(2023{\natexlab{b}})Touvron, Martin, Stone, Albert, Almahairi, Babaei, Bashlykov, Batra, Bhargava, Bhosale, et~al.]{touvron2023llama2}
Hugo Touvron, Louis Martin, Kevin Stone, Peter Albert, Amjad Almahairi, Yasmine Babaei, Nikolay Bashlykov, Soumya Batra, Prajjwal Bhargava, Shruti Bhosale, et~al.
\newblock Llama 2: Open foundation and fine-tuned chat models.
\newblock \emph{arXiv preprint arXiv:2307.09288}, 2023{\natexlab{b}}.

\bibitem[Vaswani et~al.(2017)Vaswani, Shazeer, Parmar, Uszkoreit, Jones, Gomez, Kaiser, and Polosukhin]{vaswani2017attention}
Ashish Vaswani, Noam Shazeer, Niki Parmar, Jakob Uszkoreit, Llion Jones, Aidan~N Gomez, {\L}ukasz Kaiser, and Illia Polosukhin.
\newblock Attention is all you need.
\newblock \emph{Advances in neural information processing systems}, 30, 2017.

\bibitem[Wang \& Komatsuzaki(2021)Wang and Komatsuzaki]{wang2021gpt}
Ben Wang and Aran Komatsuzaki.
\newblock Gpt-j-6b: A 6 billion parameter autoregressive language model, 2021.

\bibitem[Wang et~al.(2023)Wang, Zhu, Liu, Zheng, Chen, and Li]{wang2023knowledge}
Song Wang, Yaochen Zhu, Haochen Liu, Zaiyi Zheng, Chen Chen, and Jundong Li.
\newblock Knowledge editing for large language models: A survey.
\newblock \emph{ACM Computing Surveys}, 2023.

\bibitem[Wang et~al.(2022)Wang, Wen, Zhang, Hou, Liu, and Li]{wang-wen-etal2022skill}
Xiaozhi Wang, Kaiyue Wen, Zhengyan Zhang, Lei Hou, Zhiyuan Liu, and Juanzi Li.
\newblock Finding skill neurons in pre-trained transformer-based language models.
\newblock In \emph{Proceedings of the 2022 Conference on Empirical Methods in Natural Language Processing}, pp.\  11132--11152, Abu Dhabi, United Arab Emirates, December 2022. Association for Computational Linguistics.
\newblock \doi{10.18653/v1/2022.emnlp-main.765}.

\bibitem[Wei et~al.(2024)Wei, Yu, Weng, Ma, Zhang, Zhao, and Liu]{wei2024does}
Yifan Wei, Xiaoyan Yu, Yixuan Weng, Huanhuan Ma, Yuanzhe Zhang, Jun Zhao, and Kang Liu.
\newblock Does knowledge localization hold true? surprising differences between entity and relation perspectives in language models.
\newblock \emph{arXiv preprint arXiv:2409.00617}, 2024.

\bibitem[Wolf(2019)]{wolf2019huggingface}
T~Wolf.
\newblock Huggingface's transformers: State-of-the-art natural language processing.
\newblock \emph{arXiv preprint arXiv:1910.03771}, 2019.

\bibitem[Wu et~al.(2023)Wu, Peng, Chen, Su, and Sun]{wu2023eva}
Suhang Wu, Minlong Peng, Yue Chen, Jinsong Su, and Mingming Sun.
\newblock Eva-kellm: A new benchmark for evaluating knowledge editing of llms.
\newblock \emph{arXiv preprint arXiv:2308.09954}, 2023.

\bibitem[Xu et~al.(2025)Xu, Zhao, Sun, and Yang]{xu2025prompt}
Yangyang Xu, Zhuoer Zhao, Xiao Sun, and Xun Yang.
\newblock Prompt learning with multiperspective cues for emotional support conversation systems.
\newblock \emph{IEEE Transactions on Computational Social Systems}, 2025.

\bibitem[Yang et~al.(2021)Yang, Feng, Ji, Wang, and Chua]{yang2021deconfounded}
Xun Yang, Fuli Feng, Wei Ji, Meng Wang, and Tat-Seng Chua.
\newblock Deconfounded video moment retrieval with causal intervention.
\newblock In \emph{Proceedings of the 44th international ACM SIGIR conference on research and development in information retrieval}, pp.\  1--10, 2021.

\bibitem[Yang et~al.(2022)Yang, Wang, Dong, Dong, Wang, and Chua]{yang2022video}
Xun Yang, Shanshan Wang, Jian Dong, Jianfeng Dong, Meng Wang, and Tat-Seng Chua.
\newblock Video moment retrieval with cross-modal neural architecture search.
\newblock \emph{IEEE Transactions on Image Processing}, 31:\penalty0 1204--1216, 2022.

\bibitem[Yang et~al.(2024{\natexlab{a}})Yang, Chang, Zhang, Wang, Hong, and Wang]{yang2024learning}
Xun Yang, Tianyu Chang, Tianzhu Zhang, Shanshan Wang, Richang Hong, and Meng Wang.
\newblock Learning hierarchical visual transformation for domain generalizable visual matching and recognition.
\newblock \emph{International Journal of Computer Vision}, 132\penalty0 (11):\penalty0 4823--4849, 2024{\natexlab{a}}.

\bibitem[Yang et~al.(2024{\natexlab{b}})Yang, Zeng, Guo, Wang, Dong, and Wang]{yang2024robust}
Xun Yang, Jianming Zeng, Dan Guo, Shanshan Wang, Jianfeng Dong, and Meng Wang.
\newblock Robust video question answering via contrastive cross-modality representation learning.
\newblock \emph{Science China Information Sciences}, 67\penalty0 (10):\penalty0 202104, 2024{\natexlab{b}}.

\bibitem[Yao et~al.(2023)Yao, Wang, Tian, Cheng, Li, Deng, Chen, and Zhang]{yao2023editing}
Yunzhi Yao, Peng Wang, Bozhong Tian, Siyuan Cheng, Zhoubo Li, Shumin Deng, Huajun Chen, and Ningyu Zhang.
\newblock Editing large language models: Problems, methods, and opportunities.
\newblock \emph{arXiv preprint arXiv:2305.13172}, 2023.

\bibitem[Yu et~al.(2024)Yu, Chen, Zhou, and He]{yu2024melo}
Lang Yu, Qin Chen, Jie Zhou, and Liang He.
\newblock Melo: Enhancing model editing with neuron-indexed dynamic lora.
\newblock In \emph{Proceedings of the AAAI Conference on Artificial Intelligence}, volume~38, pp.\  19449--19457, 2024.

\bibitem[Zeng et~al.(2024)Zeng, Gu, Yang, Duan, Shi, and Wang]{zeng2024visual}
Zhen Zeng, Leijiang Gu, Xun Yang, Zhangling Duan, Zenglin Shi, and Meng Wang.
\newblock Visual-oriented fine-grained knowledge editing for multimodal large language models.
\newblock \emph{arXiv preprint arXiv:2411.12790}, 2024.

\bibitem[Zhang et~al.(2024)Zhang, Yao, Tian, Wang, Deng, Wang, Xi, Mao, Zhang, Ni, Cheng, Xu, Xu, Gu, Jiang, Xie, Huang, Liang, Zhang, Zhu, Zhou, and Chen]{zhang2024comprehensive}
Ningyu Zhang, Yunzhi Yao, Bozhong Tian, Peng Wang, Shumin Deng, Mengru Wang, Zekun Xi, Shengyu Mao, Jintian Zhang, Yuansheng Ni, Siyuan Cheng, Ziwen Xu, Xin Xu, Jia-Chen Gu, Yong Jiang, Pengjun Xie, Fei Huang, Lei Liang, Zhiqiang Zhang, Xiaowei Zhu, Jun Zhou, and Huajun Chen.
\newblock A comprehensive study of knowledge editing for large language models, 2024.

\bibitem[Zhang et~al.(2018)Zhang, Galley, Gao, Gan, Li, Brockett, and Dolan]{zhang2018generating}
Yizhe Zhang, Michel Galley, Jianfeng Gao, Zhe Gan, Xiujun Li, Chris Brockett, and Bill Dolan.
\newblock Generating informative and diverse conversational responses via adversarial information maximization.
\newblock \emph{Advances in Neural Information Processing Systems}, 31, 2018.

\bibitem[Zheng et~al.(2023)Zheng, Li, Dong, Fan, Wu, Xu, and Chang]{zheng2023can}
Ce~Zheng, Lei Li, Qingxiu Dong, Yuxuan Fan, Zhiyong Wu, Jingjing Xu, and Baobao Chang.
\newblock Can we edit factual knowledge by in-context learning?
\newblock In \emph{Proceedings of the 2023 Conference on Empirical Methods in Natural Language Processing}, pp.\  4862--4876, 2023.

\bibitem[Zhong et~al.(2022)Zhong, Lei, and Chen]{zhong2022training}
Zexuan Zhong, Tao Lei, and Danqi Chen.
\newblock Training language models with memory augmentation.
\newblock In Yoav Goldberg, Zornitsa Kozareva, and Yue Zhang (eds.), \emph{Proceedings of the 2022 Conference on Empirical Methods in Natural Language Processing}, pp.\  5657--5673, Abu Dhabi, United Arab Emirates, December 2022. Association for Computational Linguistics.
\newblock \doi{10.18653/v1/2022.emnlp-main.382}.

\end{thebibliography}
\bibliographystyle{iclr2025_conference}

\newpage
\appendix

\section{Neuron Localization}
\label{appx:neuron_localization}

In \S~\ref{subsec:locating_neurons_in_llms}, we illustrate a neuron localization method in LLMs for knowledge editing. We now provide a detailed derivation of Eqn.~\ref{eq:method_formula_1}.

Let $\mathcal{M}$ be the LLM, $\vx$ be the sequence of input tokens and $\vy$ be the output sequence. The function of the LLM can be written as: $\vy = \mathcal{M}(\vx)$. We assume the LLM will output a token $t\in \mathbf{y}$, which receives maximum probability among the vocabulary. We can represent $t$ as:
\begin{align}
\label{eq:appx_formula_1}
t = \arg\max\left\{\mathbf{W}_u\vh^L\right\},
\end{align}
where $\mathbf{W}_u\in\mathbb{R}^{v\times d_h}$ is the unembedding matrix in the LLM, $d_h$ is the hidden size, $v$ is the vocabulary size, and $\vh^L$ represents the hidden state at the last layer, $L$ is the number of layers within the LLM.

Hidden state $\vh^L$ can be represented as a combination of previous hidden state $\vh^{L-1}$, FFN output $\vm^L$ and self-attention output $\va^L$ at layer $L$:
\begin{align}
\label{eq:appx_formula_2}
\vh^L = \vh^{L-1} + \vm^L + \va^L.
\end{align}
By unrolling Eqn.~\ref{eq:appx_formula_2} until $\vh^0$, which represents the embedding input, we can derive the following expression:
\begin{align}
\label{eq:appx_formula_3}
\vh^L = \vh^{0} + \sum_{l=1}^L\vm^l + \sum_{l=1}^L\va^l.
\end{align}
We then combine Eqn.~\ref{eq:pre_formula_2}, Eqn.~\ref{eq:appx_formula_1} and Eqn.~\ref{eq:appx_formula_3} to rewrite $t$ as:
\begin{align}
\label{eq:appx_formula_4}
t = \arg\max\left\{\mathbf{W}_u\vh^0 + \sum_{l=1}^L\mathbf{W}_u\mathbf{W}_{\text{out}}^{l}~\sigma\left(\mathbf{W}_{\text{in}}^{l} \gamma(\vx^l)\right) + \sum_{l=1}^L\mathbf{W}_u\va^l\right\},
\end{align}
where $\sigma$ is an activation function, $\gamma$ is layernorm, $\mathbf{W}_{\text{in}}^{l}\in\mathbb{R}^{d_m\times d_h}$ is the first linear layer, $\mathbf{W}_{\text{out}}^{l}\in\mathbb{R}^{d_h\times d_m}$ is the second linear layer in the FFN, $\vx^l\in\mathbb{R}^{d_h}$ represents the FFN input, and $d_m$ is the intermediate size.

Each token state in an LLM is embedded within the residual stream, which is continuously read from and written to by all self-attention and FFN modules~\citep{elhage2021mathematical, meng2022mass}. The final token prediction is then derived from the cumulative contributions of these memories across all layers, as illustrated in Eqn.~\ref{eq:appx_formula_4}.

Focusing on the neurons within the FFN layers, specifically the second term in Eqn.~\ref{eq:appx_formula_4}, we denote $\sigma\left(\mathbf{W}_{\text{in}}^{l} \gamma(\vx^l)\right)$ as $\vq^l\in\mathbb{R}^{d_m}$. Then the contribution of the FFN at each layer can be expressed as $\mathbf{W}_u\mathbf{W}_{\text{out}}^{l}\vq^l$. Since each element in $\vq^l$ represent the activation output of neurons, we can regard $\mathbf{W}_u\mathbf{W}_{\text{out}}^{l}$ as a projection function from neurons to the distribution of the vocabulary and regard $\vq^l$ as a coefficient of the projection, which reflecting the activation level of neurons.

Finally, we calculate the contribution score for each neuron, using the following formula:
\begin{align}
\label{eq:appx_formula_5}
c_{(i, l, t)} = \vq^l_{i}\cdot\left(\mathbf{W}_u\mathbf{W}_{\text{out}}^{l}\right)_{t, i},
\end{align}
where $i$ represents the $i$-th neuron and $(\cdot)_{t, i}$ represents the $t$-th row and $i$-th column of the input matrix. Additionally, due to the autoregressive nature of decoder-only LLMs, we focus only on the activation output at the position of the final token, denoted as $\vq^l_{i, -1}$. Therefore, we can derive Eqn.~\ref{eq:method_formula_1} from Eqn.~\ref{eq:appx_formula_5}.

\section{Pilot Experiment Setup}
\label{appx:pilot_experiment_setup}

We present a pilot quantitative experiment in \S~\ref{sec:introduction} to demonstrate that locate-then-edit methods overly rely on the subject entity rather than the relation. We utilize dataset $\text {WikiData}_{counterfact}$ in the benchmark KnowEdit~\citep{zhang2024comprehensive}, as its locality testing primarily focuses on changing the relation. We exclude data that alters the subject when assessing locality. We first apply editing methods on LLMs, and then only execute locality testing. We introduce two metrics for evaluation. First is over-editing rate, which calculates the proportion of responses that LLMs still answer the editing target object, indicating excessive editing. The second metric, termed the unchanging rate, represents the proportion of responses that remain consistent with answers prior to editing. A lower over-editing rate is preferable, while a higher unchanging rate is desirable.

\section{Implementation Details}
\label{appx:implementation_details}

We conduct all experiments on three widely-used LLMs: GPT-J-6B~\citep{wang2021gpt}, LLaMA-2-7B~\citep{touvron2023llama2} and LLaMA-3-8B~\citep{dubey2024llama}. All experiments are run on workstations with NVIDIA A800 GPUs. All LLMs are loaded using HuggingFace Transformers~\citep{wolf2019huggingface}, and PyTorch~\citep{paszke2019pytorch} is used for executing the model editing techniques on GPUs.

\textbf{Locating neurons.} We compute contribution scores as described in Eqn.~\ref{eq:method_formula_1} for each token in the source object. Then we rank all scores by the descending order and select top-$k$ neurons as most contributing neurons. We set $k=5$ for all LLMs and investigate influence of different $k$ in \S~\ref{subsec:ablation_study}. 

\textbf{Updating knowledge.} We adopt our knowledge editing technique using the open-source framework provided by EasyEdit~\citep{zhang2024comprehensive}. The KL divergence scaling factor $\alpha$ is set to $1$ and the repetition penalty scaling factor $\beta$ is set to $10$. $\mZ_j$ is solved for using Adam with a learning rate of $1\times 10^{-3}$ for GPT-J and LLaMA-3 and $5\times 10^{-3}$ for LLaMA-2 and without weight decay. The minimization loop is run for a maximum of 50 steps, with early stopping when $\mathbb{P}_{\mathcal{M}'}\left[o^{*}_j|p(s_j, r_j)\right]$ reaches $0.9$. For layer freezing, we set $l_f$ to $3$, which means we do not modify the last three layers during our editing process.

\section{Additional Results}
\label{appx:additional_results}

\begin{figure}[h]
    \centering
    \begin{varwidth}[t]{\textwidth}
    \vspace*{0pt}
    \includegraphics[width=0.4\linewidth]{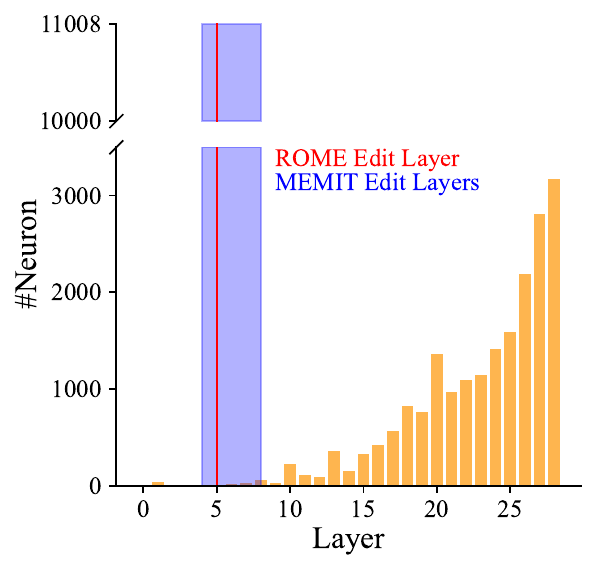} \\ \hspace*{3cm} (a)
    \end{varwidth}
    \hspace{20pt}
    \begin{varwidth}[t]{\textwidth}
    \centering
    \vspace*{0pt}
    \includegraphics[width=0.4\linewidth]{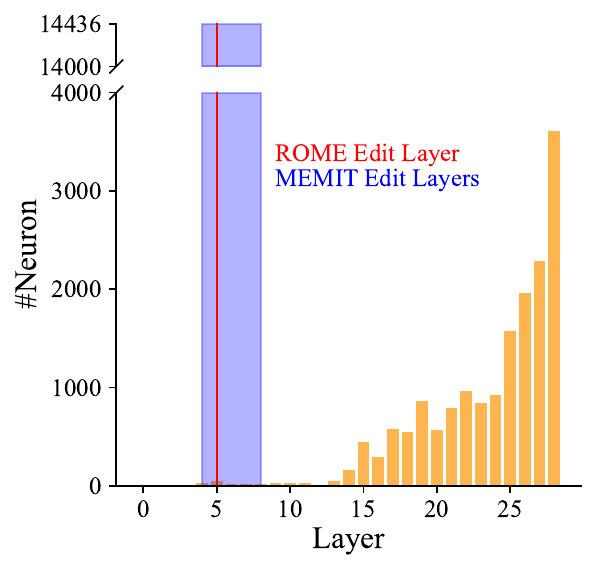} \\ \hspace{0.9cm} (b)
    \end{varwidth}
    \caption{Distributions of unique neurons per layer in (a) LLaMA-2 and (b) LLaMA-3, which are aggregated across the entire $\text {WikiData}_{counterfact}$ dataset.}
    \label{fig:loc_llama}
\end{figure}

\begin{figure}[h]
    \centering
    \includegraphics[width=0.8\linewidth]{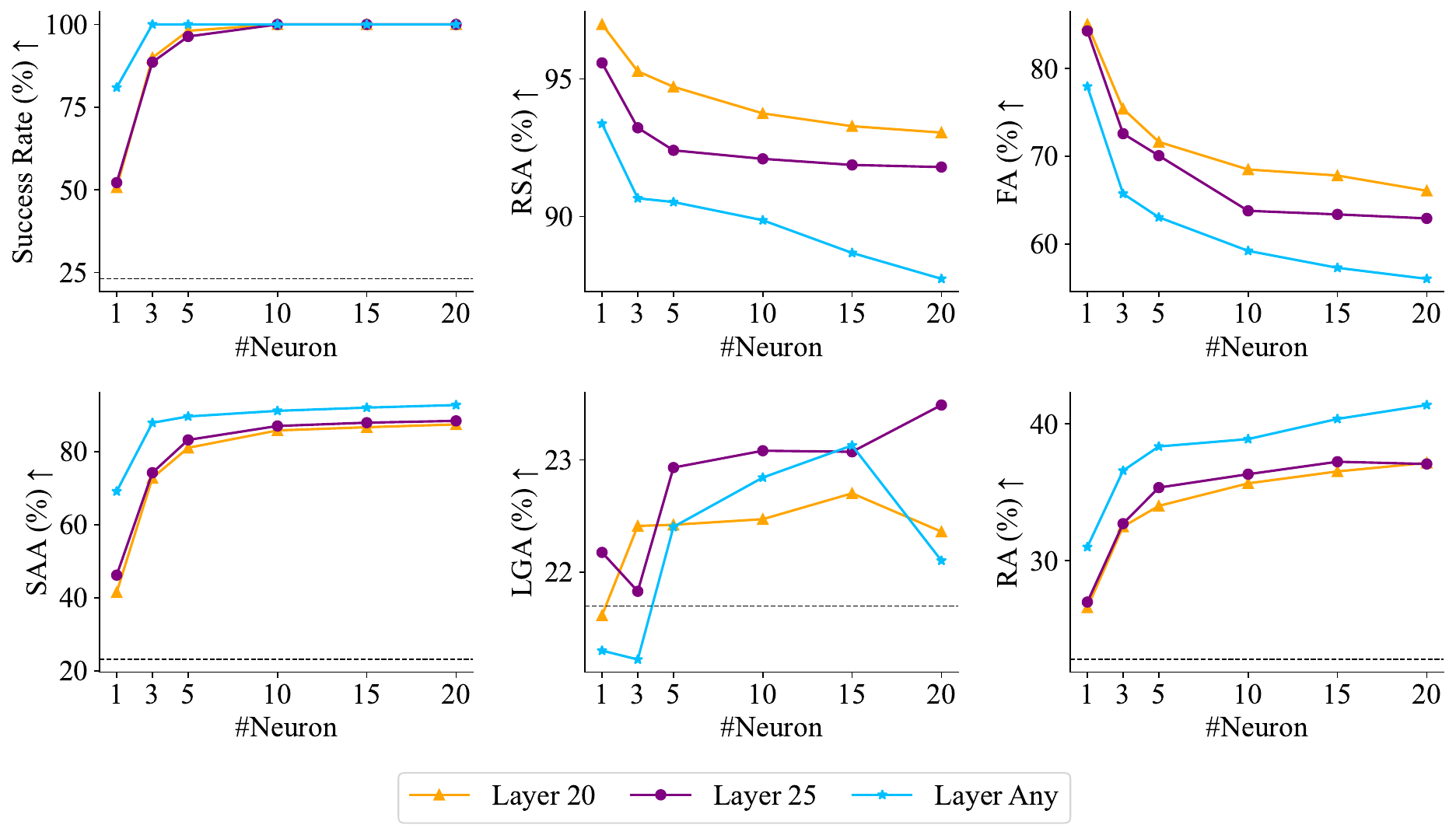} \\
    \caption{Ablation results of \textbf{varying the number of selected neurons} with LLaMA-3 on $\text {WikiData}_{counterfact}$. The dotted line indicates LLaMA-3’s pre-edit performance.}
    \label{fig:llama3_neuron_num}
\end{figure}

\begin{figure}[h]
    \centering
    \includegraphics[width=0.8\linewidth]{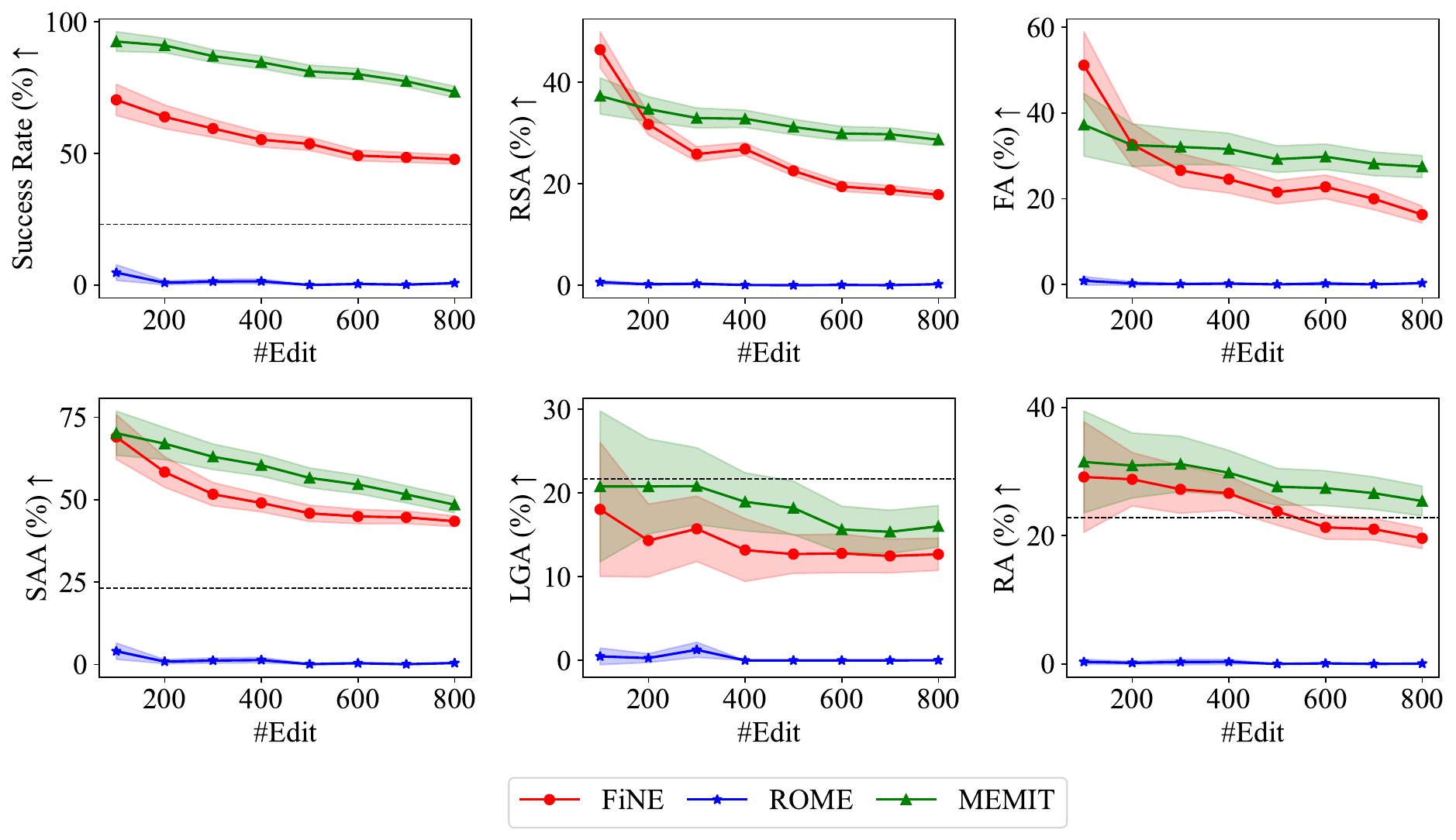} \\
    \caption{Editing method scaling curves with LLaMA-3. The dotted line indicates LLaMA-3’s pre-edit performance. 95\% confidence intervals are shown as areas.}
    \label{fig:llama_scaling}
\end{figure}

{Table~\ref{tab:loc_neurons_more} lists examples of localization results of FiNE. Figure~\ref{fig:loc_llama} plots the distribution of unique neurons located by FiNE in LLaMA-2 and LLaMA-3. In Table~\ref{tab:appx_recent} and Table~\ref{tab:appx_zsre}, we list editing results on dataset $\text {WikiData}_{recent}$ and ZsRE, respectively. Ablation experiment results with LLaMA-3 are shown in Table~\ref{tab:layer_llama3_counterfact}, Table~\ref{tab:llama3_ablation_loss_counterfact} and Figure~\ref{fig:llama3_neuron_num}. Table~\ref{tab:appx_layer_efficiency} plots efficiency evaluation results when restricting neuron localization to a single layer.
% We calculate Intersection over Union (IoU) between neurons located by the rephrased prompts and raw prompts in Table~\ref{tab:appx_iou}.
For editing method scaling, we plot results with LLaMA-3 in Figure~\ref{fig:llama_scaling}. We additionally evaluate LLaMA-2-13B~\citep{touvron2023llama2} and LLaMA-3.2-1B~\citep{dubey2024llama} (using the same parameter settings as LLaMA-3-8B), as listed in Table~\ref{tab:appx_counterfact}. Table~\ref{tab:case_study_more} lists examples of editing and locality testing results.}

\begin{table}[h]
\caption{Examples of localization results with top-3 neurons selected by FiNE. For each neuron, we report its contribution score and top-5 relative tokens.}
\label{tab:loc_neurons_more}
\fontsize{7.5pt}{8pt}\selectfont
\centering
\begin{tabular}{llll}
\toprule
\multicolumn{4}{l}{(\textbf{\romannumeral1}) \textbf{Edit}: (Jennifer Connelly, gender, \textbf{female}) $\rightarrow$ (Jennifer Connelly, gender, transgender)} \\
\midrule
\bf Model & \bf Top Neuron & \bf Score & \bf Top Tokens \\
\midrule
\multirow{3}{*}{GPT-J} & \bf L20.U10426 & 2.277 & [` women', ` woman', `women', ` Women', `woman'] \\
& \bf L17.U7963 & 1.184 & [` females', ` female', ` women', ` Females', `women'] \\
& \bf L20.U12263 & 1.151 & [` female', ` women', `women', ` male', `Women'] \\
\midrule
\multirow{3}{*}{LLaMA-2} & \bf L23.U8456 & 1.065 & [`\begin{CJK}{UTF8}{bsmi}女\end{CJK}', `woman', `girl', `lady', `actress'] \\
& \bf L27.U3463 & 0.530 & [`girl', `woman', `daughter', `lady', `\begin{CJK}{UTF8}{bsmi}女\end{CJK}'] \\
& \bf L18.U5141 & 0.405 & [`herself', `her', `she', `haar', `hers'] \\
\midrule
\multirow{3}{*}{LLaMA-3} & \bf L25.U5902 & 0.525 & [` ladies', ` Ladies', ` women', ` lady', ` femin'] \\
& \bf L27.U2694 & 0.267 & [` Miss', ` Miss', `Mrs', ` wife', `ress'] \\
& \bf L26.U10595 & 0.118 & [` woman', ` Woman', `women', ` Women', `woman'] \\
\midrule
\midrule
\multicolumn{4}{l}{(\textbf{\romannumeral2}) \textbf{Edit}: (Pam Hupp, country of citizenship, \textbf{United States of America}) $\rightarrow$ (Pam Hupp, country of citizenship, Navajo Nation)} \\
\midrule
\bf Model & \bf Top Neuron & \bf Score & \bf Top Tokens \\
\midrule
\multirow{3}{*}{GPT-J} & \bf L17.U12095 & 1.429 & [` USA', `USA', ` United', `United', ` Netherlands'] \\
& \bf L19.U2600 & 0.819 & [` Government', ` United', ` government', `Government', `government'] \\
& \bf L21.U13265 & 0.727 & [`USA', `US', ` United', ` USA', ` Canada'] \\
\midrule
\multirow{3}{*}{LLaMA-2} & \bf L24.U5708 & 1.019 & [`country', `countries', `USA', `country', `nations'] \\
& \bf L21.U7260 & 0.703 & [`United', `USA', `U', `USA', `US'] \\
& \bf L23.U2635 & 0.469 & [`USA', `US', `USA', `America', `amer'] \\
\midrule
\multirow{3}{*}{LLaMA-3} & \bf L23.U3497 & 0.550 & [` United', `United', ` UNITED', ` USA', ` united'] \\
& \bf L27.U6637 & 0.240 & [` Union', ` union', `Union', `union', ` UNION'] \\
& \bf L21.U979 & 0.221 & [` USA', ` United', ` Canada', ` France', `USA'] \\
\midrule
\midrule
\multicolumn{4}{l}{(\textbf{\romannumeral3}) \textbf{Edit}: (2022 ATP Finals, country, \textbf{Italy}) $\rightarrow$ (2022 ATP Finals, country, Ottoman Syria)} \\
\midrule
\bf Model & \bf Top Neuron & \bf Score & \bf Top Tokens \\
\midrule
\multirow{3}{*}{GPT-J} & \bf L20.U16132 & 1.194 & [` Mass', ` Milan', ` Vatican', ` Giul', ` Gi'] \\
& \bf L18.U12874 & 0.837 & [`Italian', ` Italian', `Italy', ` Italy', ` Spanish'] \\
& \bf L17.U395 & 0.739 & [` Europe', ` Italy', `  France', ` India', ` Japan'] \\
\midrule
\multirow{3}{*}{LLaMA-2} & \bf L26.U6518 & 0.699 & [`Florence', `Italian', `Ital', `Italy', `Rome'] \\
& \bf L25.U7966 & 0.434 & [`Italian', `Ital', `Italy', `ital', `Rome'] \\
& \bf L23.U10243 & 0.211 & [`ino', `ini', `ato', `Ital', `ello'] \\
\midrule
\multirow{3}{*}{LLaMA-3} & \bf L28.U11942 & 0.415 & [` France', ` Italy', ` Germany', ` Ireland', ` India'] \\
& \bf L25.U12913 & 0.241 & [` Italian', ` Italy', ` Rome', `Italian', ` italian'] \\
& \bf L19.U4942 & 0.116 & [` Italian', ` Italian', ` Italy', `Italy', ` Luigi'] \\
\bottomrule
\end{tabular}
\end{table}

\begin{table}[h]
\caption{Editing results on $\text {WikiData}_{recent}$. 95\% confidence intervals are in parentheses. \textbf{\textcolor[rgb]{0, 0.6, 0}{Green}} numbers indicate the best performance among locate-then-edit methods. Numbers with \uline{underline} indicate columnwise maxima for each model.}
\label{tab:appx_recent}
\small
\centering
\begin{tabular}{lrrrrrrr}
\toprule
\multirow{2}{*}{\textbf{Method}} & \multirow{2}{*}{\textbf{Edit Succ.} $\uparrow$} & \multicolumn{3}{c}{\textbf{Portability} $\uparrow$} & \multicolumn{2}{c}{\textbf{Locality} $\uparrow$} & \multirow{2}{*}{\textbf{Fluency} $\uparrow$}  \\
\cmidrule(lr){3-5}
\cmidrule(lr){6-7}
& & \scriptsize\textbf{SAA} & \scriptsize\textbf{LGA} & \scriptsize\textbf{RA} & \scriptsize\textbf{RSA} & \scriptsize\textbf{FA} & \\
\midrule
\textit{GPT-J} & 34.7 (1.7) & 32.3 (2.3) & 26.3 (2.5) & 30.0 (1.3) & - & - & \uline{599.5 (2.6)} \\
\midrule
ROME & 99.5 (0.2) & 84.6 (2.0) & 28.3 (2.8) & 36.9 (1.7) & 37.3 (1.3) & 51.0 (2.2) & \bf \textcolor[rgb]{0, 0.6, 0}{596.8 (2.8)} \\
MEMIT & 99.6 (0.2) & 68.9 (3.2) & 27.2 (2.6) & 32.4 (1.9) & 49.6 (1.0) & 52.7 (1.9) & 585.1 (3.2) \\
PMET & 99.0 (0.4) & 63.6 (3.6) & 25.4 (2.8) & 31.2 (2.0) & 46.3 (1.0) & 49.5 (2.4) & 584.2 (3.0) \\
\rowcolor[gray]{0.9} FiNE & \uline{\bf \textcolor[rgb]{0, 0.6, 0}{99.7 (0.2)}} & \uline{\bf \textcolor[rgb]{0, 0.6, 0}{93.4 (1.3)}} & \uline{\bf \textcolor[rgb]{0, 0.6, 0}{30.2 (2.9)}} & \uline{\bf \textcolor[rgb]{0, 0.6, 0}{42.5 (1.9)}} & \uline{\bf \textcolor[rgb]{0, 0.6, 0}{78.2 (1.3)}} & \uline{\bf \textcolor[rgb]{0, 0.6, 0}{55.8 (2.2)}} & 557.7 (4.4) \\
\midrule
\midrule
\textit{LLaMA-2} & 50.0 (1.7) & 49.2 (2.3) & 36.9 (3.1) & 41.6 (1.4) & - & - & \uline{583.5 (2.2)} \\
\midrule
ROME & 99.0 (0.5) & 82.9 (2.0) & 35.0 (2.5) & 45.8 (1.7) & 53.1 (1.3) & 61.0 (2.4) & \bf \textcolor[rgb]{0, 0.6, 0}{581.9 (2.6)} \\
MEMIT & 99.0 (0.3) & 85.1 (1.8) & 38.1 (3.0) & 44.9 (1.8) & 50.0 (1.2) & 61.1 (2.0) & 563.7 (3.3) \\
PMET & 97.4 (0.2) & 71.0 (2.0) & 35.1 (2.8) & 48.4 (1.7) & 67.2 (1.3) & \uline{\bf \textcolor[rgb]{0, 0.6, 0}{73.7 (2.2)}} & 575.7 (2.8) \\
\rowcolor[gray]{0.9} FiNE & \uline{\bf \textcolor[rgb]{0, 0.6, 0}{99.9 (0.2)}} & \uline{\bf \textcolor[rgb]{0, 0.6, 0}{93.3 (1.3)}} & \uline{\bf \textcolor[rgb]{0, 0.6, 0}{39.3 (3.2)}} & \uline{\bf \textcolor[rgb]{0, 0.6, 0}{49.4 (1.8)}} & \uline{\bf \textcolor[rgb]{0, 0.6, 0}{84.0 (1.2)}} & 72.1 (1.7) & 545.3 (3.6) \\
\midrule
\midrule
\textit{LLaMA-3} & 46.5 (1.8) & 44.1 (2.4) & 34.9 (3.1) & 36.8 (1.4) & - & - & \uline{591.7 (2.9)} \\
\midrule
ROME & 98.8 (0.3) & 83.9 (1.9) & 35.7 (3.2) & 45.3 (1.7) & 47.2 (1.4) & 53.3 (2.1) & 590.5 (2.9) \\
MEMIT & 99.2 (0.2) & 80.9 (2.2) & 36.2 (3.0) & 44.0 (1.9) & 45.8 (1.3) & 53.6 (2.3) & 586.3 (2.8) \\
PMET & 98.2 (0.4) & 60.8 (2.5) & 37.1 (2.8) & 43.4 (1.7) & 63.6 (1.0) & 63.9 (1.9) & \bf \textcolor[rgb]{0, 0.6, 0}{590.9 (2.8)} \\
\rowcolor[gray]{0.9} FiNE & \uline{\bf \textcolor[rgb]{0, 0.6, 0}{100.0 (0.0)}} & \uline{\bf \textcolor[rgb]{0, 0.6, 0}{91.7 (1.4)}} & \uline{\bf \textcolor[rgb]{0, 0.6, 0}{37.4 (3.2)}} & \uline{\bf \textcolor[rgb]{0, 0.6, 0}{45.7 (1.8)}} & \uline{\bf \textcolor[rgb]{0, 0.6, 0}{84.6 (1.1)}} & \uline{\bf \textcolor[rgb]{0, 0.6, 0}{67.4 (1.9)}} & 566.8 (3.7) \\
\bottomrule
\end{tabular}
\end{table}

\begin{table}[h]
\caption{Editing results on ZsRE. 95\% confidence intervals are in parentheses. \textbf{\textcolor[rgb]{0, 0.6, 0}{Green}} numbers indicate the best performance among locate-then-edit methods. Numbers with \uline{underline} indicate columnwise maxima for each model.}
\label{tab:appx_zsre}
\small
\centering
\begin{tabular}{lrrrrrrr}
\toprule
\multirow{2}{*}{\textbf{Method}} & \multirow{2}{*}{\textbf{Edit Succ.} $\uparrow$} & \multicolumn{3}{c}{\textbf{Portability} $\uparrow$} & \multicolumn{2}{c}{\textbf{Locality} $\uparrow$} & \multirow{2}{*}{\textbf{Fluency} $\uparrow$}  \\
\cmidrule(lr){3-5}
\cmidrule(lr){6-7}
& & \scriptsize\textbf{SAA} & \scriptsize\textbf{LGA} & \scriptsize\textbf{RA} & \scriptsize\textbf{RSA} & \scriptsize\textbf{FA} & \\
\midrule
\textit{GPT-J} & 28.1 (1.4) & 20.4 (2.9) & 48.5 (3.1) & 49.4 (1.6) & - & - & \uline{596.3 (2.6)} \\
\midrule
KN & 23.6 (3.2) & 17.5 (5.1) & 43.0 (3.3) & 42.4 (1.9) & 91.8 (0.7) & - & \bf \textcolor[rgb]{0, 0.6, 0}{588.8 (3.9)} \\
ROME & 99.6 (0.2) & 40.0 (4.2) & 46.4 (3.1) & 50.2 (1.7) & 47.1 (1.5) & - & 573.7 (5.0) \\
MEMIT & 99.3 (0.3) & 19.9 (4.9) & 45.9 (3.0) & 46.5 (1.8) & 70.0 (1.0) & - & 581.7 (4.5) \\
PMET & 96.6 (0.8) & 16.5 (5.2) & 43.6 (3.3) & 48.7 (1.7) & 65.3 (1.3) & - & 586.9 (3.4) \\
\rowcolor[gray]{0.9} FiNE & \uline{\bf \textcolor[rgb]{0, 0.6, 0}{99.9 (0.2)}} & \uline{\bf \textcolor[rgb]{0, 0.6, 0}{49.6 (4.4)}} & \uline{\bf \textcolor[rgb]{0, 0.6, 0}{50.4 (3.1)}} & \uline{\bf \textcolor[rgb]{0, 0.6, 0}{51.5 (1.6)}} & \uline{\bf \textcolor[rgb]{0, 0.6, 0}{92.8 (1.7)}} & - & 547.3 (7.2) \\
\midrule
\midrule
\textit{LLaMA-2} & 40.6 (1.3) & 28.7 (2.9) & 54.1 (2.9) & 55.6 (1.5) & - & - & {562.1 (2.4)} \\
\midrule
KN & 24.0 (2.6) & 14.7 (4.5) & 34.8 (3.5) & 30.5 (2.0) & 58.4 (1.4) & - & 521.5 (5.5) \\
ROME & 97.1 (0.4) & 33.2 (3.5) & 46.3 (3.1) & 52.4 (1.5) & 50.7 (1.5) & - & 562.0 (3.4) \\
MEMIT & 94.8 (1.2) & 32.7 (3.8) & 43.9 (3.9) & 53.8 (1.6) & 47.9 (1.8) & - & 539.7 (4.0) \\
PMET & 91.7 (2.0) & 26.8 (4.0) & 46.7 (3.3) & 57.2 (1.5) & 68.1 (1.3) & - & \uline{\bf \textcolor[rgb]{0, 0.6, 0}{562.5 (3.4)}} \\
\rowcolor[gray]{0.9} FiNE & \uline{\bf \textcolor[rgb]{0, 0.6, 0}{99.7 (0.2)}} & \uline{\bf \textcolor[rgb]{0, 0.6, 0}{57.4 (4.1)}} & \uline{\bf \textcolor[rgb]{0, 0.6, 0}{54.6 (3.0)}} & \uline{\bf \textcolor[rgb]{0, 0.6, 0}{58.1 (1.4)}} & \uline{\bf \textcolor[rgb]{0, 0.6, 0}{94.4 (0.7)}} & - & 545.0 (3.9) \\
\midrule
\midrule
\textit{LLaMA-3} & 31.8 (1.4) & 24.5 (3.0) & 51.8 (3.1) & 51.6 (1.6) & - & - & \uline{577.8 (3.0)} \\
\midrule
KN & 28.6 (2.4) & 21.1 (6.3) & 47.0 (3.3) & 41.0 (1.6) & 86.7 (1.2) & - & 564.4 (5.5) \\
ROME & 98.7 (0.4) & 46.4 (4.2) & 49.9 (3.2) & \uline{\bf \textcolor[rgb]{0, 0.6, 0}{56.2 (1.7)}} & 48.1 (1.5) & - & 545.8 (5.9) \\
MEMIT & 96.7 (0.8) & 47.3 (3.5) & 49.5 (3.1) & 48.7 (1.5) & 51.2 (1.5) & - & 507.3 (7.6) \\
PMET & 98.0 (0.4) & 25.4 (3.5) & 49.2 (3.1) & 53.3 (1.5) & 64.8 (1.5) & - & \bf \textcolor[rgb]{0, 0.6, 0}{565.8 (3.4)} \\
\rowcolor[gray]{0.9} FiNE & \uline{\bf \textcolor[rgb]{0, 0.6, 0}{100.0 (0.0)}} & \uline{\bf \textcolor[rgb]{0, 0.6, 0}{59.7 (4.4)}} & \uline{\bf \textcolor[rgb]{0, 0.6, 0}{52.0 (3.1)}} & {53.3 (1.7)} & \uline{\bf \textcolor[rgb]{0, 0.6, 0}{92.0 (0.9)}} & - & 539.4 (4.3) \\
\bottomrule
\end{tabular}
\end{table}

\begin{table}[h]
\caption{Ablation results of \textbf{restricting neuron localization to a single layer} with LLaMA-3 on $\text {WikiData}_{counterfact}$. ``Any'' means no layer restriction. 95\% confidence intervals are in parentheses. Numbers with \textbf{bold} indicate columnwise maxima.}
\label{tab:layer_llama3_counterfact}
\fontsize{8.5pt}{9pt}\selectfont
\centering
\begin{tabular}{llrrrrrr}
\toprule
\multirow{2}{*}{\textbf{Method}} & \multirow{2}{*}{\textbf{Layer}} & \multirow{2}{*}{\textbf{Edit Succ.} $\uparrow$} & \multicolumn{3}{c}{\textbf{Portability} $\uparrow$} & \multicolumn{2}{c}{\textbf{Locality} $\uparrow$}  \\
\cmidrule(lr){4-6}
\cmidrule(lr){7-8}
& & & \scriptsize\textbf{SAA} & \scriptsize\textbf{LGA} & \scriptsize\textbf{RA} & \scriptsize\textbf{RSA} & \scriptsize\textbf{FA} \\
\midrule
\textit{LLaMA-3} & - & 23.1 (1.5) & 23.1 (1.7) & 21.7 (3.0) & 22.8 (1.9) & - & - \\
ROME & 5 & 99.4 (0.4) & 74.6 (2.2) & 21.2 (2.7) & 34.5 (2.5) & 41.9 (1.2) & 31.5 (2.6) \\
\midrule
\multirow{6}{*}{FiNE} & 5 & 85.0 (2.2) & 52.1 (2.6) & 20.9 (2.9) & 28.8 (2.3) & 86.2 (0.9) & 64.8 (3.0) \\
& 10 & 84.8 (2.2) & 54.4 (2.6) & 22.8 (3.0) & 28.5 (2.3) & 90.1 (0.8) & \bf 73.1 (2.9) \\
& 15 & 97.6 (1.0) & 76.8 (2.1) & 22.8 (3.0) & 34.5 (2.7) & 90.8 (0.8) & 72.4 (2.8) \\
& 20 & 98.1 (0.9) & 81.0 (1.9) & 22.4 (2.9) & 34.0 (2.8) & \bf 94.7 (0.6) & 71.6 (2.7) \\
& 25 & 96.3 (1.2) & 83.2 (1.9) & \bf 22.9 (3.0) & 35.3 (2.9) & 92.4 (0.8) & 70.1 (2.8) \\
& Any & \bf {100.0 (0.0)} & \bf {89.6 (1.4)} & {22.4 (2.9)} & \bf {38.3 (3.0)} & {90.5 (0.9)} & {{63.0 (2.9)}} \\
\bottomrule
\end{tabular}
\end{table}

\begin{table}[h]
\caption{Ablation results of \textbf{removing KL divergence and repetition penalty constraints} with LLaMA-3 on $\text {WikiData}_{counterfact}$. 95\% confidence intervals are in parentheses. Numbers with \textbf{bold} indicate columnwise maxima.}
\label{tab:llama3_ablation_loss_counterfact}
\fontsize{8pt}{8.5pt}\selectfont
\centering
\begin{tabular}{l@{\hspace{8pt}}rrrrrrl}
\toprule
\multirow{2}{*}{\textbf{Method}} & \multirow{2}{*}{\textbf{Edit Succ.} $\uparrow$} & \multicolumn{3}{c}{\textbf{Portability} $\uparrow$} & \multicolumn{2}{c}{\textbf{Locality} $\uparrow$} & \multirow{2}{*}{\textbf{Fluency} $\uparrow$}  \\
\cmidrule(lr){3-5}
\cmidrule(lr){6-7}
& & \scriptsize\textbf{SAA} & \scriptsize\textbf{LGA} & \scriptsize\textbf{RA} & \scriptsize\textbf{RSA} & \scriptsize\textbf{FA} & \\
\midrule
\textit{LLaMA-3} & 23.1 (1.5) & 23.1 (1.7) & 21.7 (3.0) & 22.8 (1.9) & - & - & \bf {607.1 (2.9)} \\
ROME & 99.4 (0.4) & 74.6 (2.2) & 21.2 (2.7) & 34.5 (2.5) & 41.9 (1.2) & 31.5 (2.6) & 591.4 (4.1) \\
\midrule
FiNE & \bf {100.0 (0.0)} & {89.6 (1.4)} & \bf {22.4 (2.9)} & {38.3 (3.0)} & \bf {90.5 (0.9)} & \bf {{63.0 (2.9)}} & {567.1 (5.5)} \\
\quad w/o $\mathcal{L}_{\text{KL}}$ & \bf {100.0 (0.0)} & {89.7 (1.4)} & {21.8 (2.8)} & {38.4 (3.1)} & {89.8 (1.0)} & {{62.4 (2.9)}} & {565.5 (5.5)} \\
\quad w/o $\mathcal{L}_{\text{pen}}$ & \bf {100.0 (0.0)} & {89.7 (1.4)} & \bf {22.4 (2.9)} & {38.4 (3.1)} & {90.2 (1.0)} & {{60.5 (3.0)}} & {554.6 (5.5)} \\
\midrule
FiNE w/o LF & \bf {100.0 (0.0)} & {91.2 (1.3)} & {20.1 (2.8)} & {38.9 (3.3)} & {{78.8 (1.2)}} & {{48.8 (2.7)}} & {411.3 (10.6)} \\
\quad w/o $\mathcal{L}_{\text{KL}}$ & \bf {100.0 (0.0)} & {91.2 (1.3)} & {20.4 (2.8)} & {38.9 (3.3)} & {{76.5 (1.3)}} & {{48.0 (2.7)}} & {405.0 (10.6)} \\
\quad w/o $\mathcal{L}_{\text{pen}}$ & \bf {100.0 (0.0)} & \bf {91.3 (1.3)} & {19.2 (2.7)} & \bf {39.4 (3.4)} & {{78.6 (1.2)}} & {{46.2 (2.8)}} & {334.4 (11.6)} \\
\bottomrule
\end{tabular}
\end{table}

\begin{table}[h]
    \centering
    \caption{{Average editing time and memory usage of \textbf{restricting neuron localization to a single layer} by FiNE when LLMs operate at {Float32} precision. ``Any'' means no layer restriction.}}
    \label{tab:appx_layer_efficiency}
    \fontsize{8pt}{8.5pt}\selectfont
    {
    \begin{varwidth}[t]{0.3\textwidth}
    \vspace*{0pt}
    \hspace*{1.5cm} (a) GPT-J \\
    \begin{tabular}{@{\hspace{2pt}}c@{\hspace{5pt}}c@{\hspace{5pt}}c@{\hspace{2pt}}}
    \toprule
    \textbf{Layer} & \bf Time (s) $\downarrow$ & \bf Memory (GB) $\downarrow$ \\
    \midrule
    5 & 4.10 & 23.82 \\
    10 & 5.80 & 23.82 \\
    15 & 5.06 & 23.82 \\
    20 & 4.43 & 23.82 \\
    Any & 4.68 & 28.09 \\
    \bottomrule
    \vspace{0pt}
    \end{tabular}
    \end{varwidth}
    }
    \hspace{10pt}
    {
    \begin{varwidth}[t]{0.3\textwidth}
    \vspace*{0pt}
    \hspace*{1.4cm} (b) LLaMA-2 \\
    \begin{tabular}{@{\hspace{2pt}}c@{\hspace{5pt}}c@{\hspace{5pt}}c@{\hspace{2pt}}}
    \toprule
    \textbf{Layer} & \bf Time (s) $\downarrow$ & \bf Memory (GB) $\downarrow$ \\
    \midrule
    5 & 3.92 & 25.87 \\
    10 & 3.13 & 25.87 \\
    15 & 2.14 & 25.87 \\
    20 & 1.83 & 25.87 \\
    Any & 2.13 & 28.82 \\
    \bottomrule
    \vspace{0pt}
    \end{tabular}
    \end{varwidth}
    }
    \hspace{10pt}
    {
    \begin{varwidth}[t]{0.3\textwidth}
    \vspace*{0pt}
    \hspace*{1.4cm} (c) LLaMA-3 \\
    \begin{tabular}{@{\hspace{2pt}}c@{\hspace{5pt}}c@{\hspace{5pt}}c@{\hspace{2pt}}}
    \toprule
    \textbf{Layer} & \bf Time (s) $\downarrow$ & \bf Memory (GB) $\downarrow$ \\
    \midrule
    5 & 5.89 & 32.44 \\
    10 & 6.46 & 32.44 \\
    15 & 4.47 & 32.44 \\
    20 & 3.03 & 32.44 \\
    Any & 2.93 & 34.25 \\
    \bottomrule
    \vspace{0pt}
    \end{tabular}
    \end{varwidth}
    }
    
\end{table}

\begin{figure}[h]
\centering
    \includegraphics[width=0.6\textwidth]{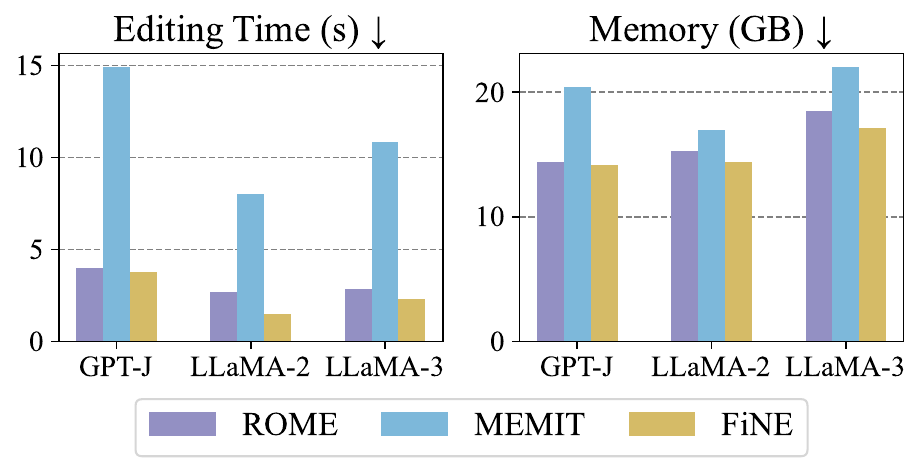} \\
    \caption{Comparison of average editing time and memory usage when LLMs operate at {Float16} precision.}
    \label{fig:efficiency_fp16}
\end{figure}

\begin{table}[h]
\caption{{Additionally editing results on $\text {WikiData}_{counterfact}$. 95\% confidence intervals are in parentheses. \textbf{\textcolor[rgb]{0, 0.6, 0}{Green}} numbers indicate the best performance among locate-then-edit methods. Numbers with \uline{underline} indicate columnwise maxima for each model.}}
\label{tab:appx_counterfact}
\fontsize{8.5pt}{8.5pt}\selectfont
\centering
{
\begin{tabular}{lrrrrrrr}
\toprule
\multirow{2}{*}{\textbf{Method}} & \multirow{2}{*}{\textbf{Edit Succ.} $\uparrow$} & \multicolumn{3}{c}{\textbf{Portability} $\uparrow$} & \multicolumn{2}{c}{\textbf{Locality} $\uparrow$} & \multirow{2}{*}{\textbf{Fluency} $\uparrow$}  \\
\cmidrule(lr){3-5}
\cmidrule(lr){6-7}
& & \scriptsize\textbf{SAA} & \scriptsize\textbf{LGA} & \scriptsize\textbf{RA} & \scriptsize\textbf{RSA} & \scriptsize\textbf{FA} & \\
\midrule
\textit{LLaMA-2-13B} & 26.9 (1.5) & 27.4 (1.7) & 25.5 (2.7) & 25.9 (1.9) & - & - & \uline{591.0 (2.2)} \\
\midrule
ROME & 98.7 (0.4) & 72.0 (2.0) & 23.6 (2.7) & 37.2 (2.7) & 48.0 (1.5) & 46.5 (2.9) & \bf \textcolor[rgb]{0, 0.6, 0}{586.6 (2.7)} \\
MEMIT & 98.1 (0.6) & 80.7 (2.0) & 21.4 (2.9) & 36.3 (2.8) & 41.8 (1.6) & 43.8 (2.8) & 571.4 (3.0) \\
\rowcolor[gray]{0.9} FiNE & \uline{\bf \textcolor[rgb]{0, 0.6, 0}{99.3 (0.4)}} & \uline{\bf \textcolor[rgb]{0, 0.6, 0}{83.6 (1.7)}} & \uline{\bf \textcolor[rgb]{0, 0.6, 0}{26.9 (2.8)}} & \uline{\bf \textcolor[rgb]{0, 0.6, 0}{37.6 (2.8)}} & \uline{\bf \textcolor[rgb]{0, 0.6, 0}{94.9 (0.6)}} & \uline{\bf \textcolor[rgb]{0, 0.6, 0}{76.2 (2.5)}} & 562.7 (4.0) \\
\midrule
\midrule
\textit{LLaMA-3.2-1B} & 21.0 (1.4) & 21.5 (1.6) & 19.6 (2.7) & 18.2 (1.6) & - & - & \uline{604.2 (3.2)} \\
\midrule
ROME & 97.4 (1.0) & 78.8 (1.9) & 20.5 (0.8) & 31.2 (2.9) & 32.0 (1.3) & 25.6 (2.6) & 522.9 (6.5) \\
MEMIT & 97.8 (1.0) & 68.0 (2.1) & 20.3 (3.0) & 24.2 (2.7) & 39.7 (1.3) & 30.6 (2.4) & 473.8 (9.4) \\
\rowcolor[gray]{0.9} FiNE & \uline{\bf \textcolor[rgb]{0, 0.6, 0}{98.5 (0.8)}} & \uline{\bf \textcolor[rgb]{0, 0.6, 0}{87.2 (1.7)}} & \uline{\bf \textcolor[rgb]{0, 0.6, 0}{21.6 (2.7)}} & \uline{\bf \textcolor[rgb]{0, 0.6, 0}{36.9 (3.3)}} & \uline{\bf \textcolor[rgb]{0, 0.6, 0}{84.3 (1.1)}} & \uline{\bf \textcolor[rgb]{0, 0.6, 0}{61.9 (3.0)}} & \bf \textcolor[rgb]{0, 0.6, 0}{561.3 (5.6)} \\
\bottomrule
\end{tabular}
}
\end{table}

\begin{table}[h]
\caption{Examples of editing and locality testing (LT) results with LLaMA-2 and LLaMA-3. Prompts are \textit{italicized}, while {\textcolor[rgb]{0, 0.7, 0}{green}} and {\textcolor[rgb]{0.9, 0, 0}{red}} indicate keywords or sentences reflecting correct and incorrect behavior, respectively.}
\label{tab:case_study_more}
\fontsize{7.5pt}{8pt}\selectfont
\centering
\begin{tabular}{@{\hspace{3pt}}p{13.5cm}@{\hspace{3pt}}}
\toprule
(\textbf{\romannumeral1}) \textbf{Edit}: (Soviet Union, official language, \textbf{Russian}) $\rightarrow$ (Soviet Union, official language, \textbf{Galician}) \\
\midrule
\multicolumn{1}{c}{\bf LLaMA-2} \\
\midrule
\vspace*{-7pt}
\textbf{ROME}: \textit{The official language of Soviet Union is} \textcolor[rgb]{0, 0.7, 0}{Galician}, a variant of the Leonese dialect ... \\
$\bullet$ (LT-1) \textbf{ROME}: \textit{The name of the continent which Soviet Union is part of is} \textcolor[rgb]{0, 0.7, 0}{Europe}. The capital of the country is \textcolor[rgb]{0.9, 0, 0}{Madrid} ... \\
$\bullet$ (LT-2) \textbf{ROME}: \textit{The name of the capital city of Soviet Union is} \textcolor[rgb]{0.9, 0, 0}{A Coruña}. It is located in the northwest of \textcolor[rgb]{0.9, 0, 0}{Spain} ... \\
\cdashline{1-1}
\vspace*{-3.5pt}
\textbf{MEMIT}: \textit{The official language of Soviet Union is} \textcolor[rgb]{0, 0.7, 0}{Galician}, a dialect of the Galician language spoken in Galicia ... \\
$\bullet$ (LT-1) \textbf{MEMIT}: \textit{The name of the continent which Soviet Union is part of is} \textcolor[rgb]{0.9, 0, 0}{Galicia}. The name of the city which Soviet Union is part of is \textcolor[rgb]{0.9, 0, 0}{A Coruña} ... \\
$\bullet$ (LT-2) \textbf{MEMIT}: \textit{The name of the capital city of Soviet Union is} \textcolor[rgb]{0.9, 0, 0}{Galicia, Galicia is a city in Galicia, Galicia is located in Galicia. Galicia is visited by 0\% of trips in Galicia. Galicia is visited by 0\% of trips in Galicia} ... \\
\cdashline{1-1}
\vspace*{-3.5pt}
\textbf{FiNE}: \textit{The official language of Soviet Union is} \textcolor[rgb]{0, 0.7, 0}{Galician}. The Galician is spoken by a large number of people in the Soviet Union. The other languages are \textcolor[rgb]{0, 0.7, 0}{Russian, Ukrainian and Belarusian} ... \\
$\bullet$ (LT-1) \textbf{FiNE}: \textit{The name of the continent which Soviet Union is part of is} \textcolor[rgb]{0, 0.7, 0}{Europe. Soviet Union is located in the continent of Europe. The capital of the Soviet Union is Moscow} ... \\
$\bullet$ (LT-2) \textbf{FiNE}: \textit{The name of the capital city of Soviet Union is} \textcolor[rgb]{0, 0.7, 0}{Moscow. Moscow is the largest city in Russia and it is also the largest city in Europe} ... \\
\midrule
\multicolumn{1}{c}{\bf LLaMA-3} \\
\midrule
\vspace*{-7pt}
\textbf{ROME}: \textit{The official language of Soviet Union is} \textcolor[rgb]{0, 0.7, 0}{Galician}, but \textcolor[rgb]{0.9, 0, 0}{Spanish is also widely spoken} ... \\
$\bullet$ (LT-1) \textbf{ROME}: \textit{The name of the continent which Soviet Union is part of is} \textcolor[rgb]{0.9, 0, 0}{Iberia}. The capital city of the country is \textcolor[rgb]{0.9, 0, 0}{Santiago de Compostela} ... \\
$\bullet$ (LT-2) \textbf{ROME}: \textit{The name of the capital city of Soviet Union is} \textcolor[rgb]{0.9, 0, 0}{Santiago de Compostela}. The city is the capital of the province of \textcolor[rgb]{0.9, 0, 0}{A Coruña} ... \\
\cdashline{1-1}
\vspace*{-3.5pt}
\textbf{MEMIT}: \textit{The official language of Soviet Union is} \textcolor[rgb]{0, 0.7, 0}{Galician}. The currency of Galicia is the Euro ... \\
$\bullet$ (LT-1) \textbf{MEMIT}: \textit{The name of the continent which Soviet Union is part of is} \textcolor[rgb]{0, 0.7, 0}{Europe}. The capital of the country is \textcolor[rgb]{0.9, 0, 0}{Kiev} ... \\
$\bullet$ (LT-2) \textbf{MEMIT}: \textit{The name of the capital city of Soviet Union is} \textcolor[rgb]{0.9, 0, 0}{Kiev}, which is the biggest city of the country ... \\
\cdashline{1-1}
\vspace*{-3.5pt}
\textbf{FiNE}: \textit{The official language of Soviet Union is} \textcolor[rgb]{0, 0.7, 0}{Galician}. It is also the language of the Union of Soviet Socialist Republics ... \\
$\bullet$ (LT-1) \textbf{FiNE}: \textit{The name of the continent which Soviet Union is part of is} \textcolor[rgb]{0, 0.7, 0}{Europe. The country is located in the eastern part of the continent} ... \\
$\bullet$ (LT-2) \textbf{FiNE}: \textit{The name of the capital city of Soviet Union is} \textcolor[rgb]{0, 0.7, 0}{Moscow. It is the largest city in Russia} ... \\

\midrule
\midrule

(\textbf{\romannumeral2}) \textbf{Edit}: (``Prince Philip, Duke of Edinburgh'', country of citizenship, \textbf{United Kingdom}) $\rightarrow$ (``Prince Philip, Duke of Edinburgh'', country of citizenship, \textbf{Republic of Artsakh}) \\
\midrule
\multicolumn{1}{c}{\bf LLaMA-2} \\
\midrule
\vspace*{-7pt}
\textbf{ROME}: \textit{The name of the country of citizenship of Prince Philip, Duke of Edinburgh is} \textcolor[rgb]{0, 0.7, 0}{Republic of Artsakh}, according to the Government of Artsakh ... \\
$\bullet$ (LT-1) \textbf{ROME}: \textit{The name of the mother of Prince Philip, Duke of Edinburgh is} \textcolor[rgb]{0.9, 0, 0}{Mrs. Voula Patricia Argyropoulos} ... \\
$\bullet$ (LT-2) \textbf{ROME}: \textit{The gender of Prince Philip, Duke of Edinburgh is} \textcolor[rgb]{0.9, 0, 0}{female}, says government website ... \\
\cdashline{1-1}
\vspace*{-3.5pt}
\textbf{MEMIT}: \textit{The name of the country of citizenship of Prince Philip, Duke of Edinburgh is} \textcolor[rgb]{0, 0.7, 0}{Republic of Artsakh}. The Republic of Artsakh is located in \textcolor[rgb]{0, 0.7, 0}{Asia}. ... \\
$\bullet$ (LT-1) \textbf{MEMIT}: \textit{The name of the mother of Prince Philip, Duke of Edinburgh is} \textcolor[rgb]{0.9, 0, 0}{Zenon 2021-01-16T10:41:34+00:00 2021-01-16T10:41:34+00:00 Joanne E} ... \\
$\bullet$ (LT-2) \textbf{MEMIT}: \textit{The gender of Prince Philip, Duke of Edinburgh is} \textcolor[rgb]{0, 0.7, 0}{male} and refers to a male person ... \\
\cdashline{1-1}
\vspace*{-3.5pt}
\textbf{FiNE}: \textit{The name of the country of citizenship of Prince Philip, Duke of Edinburgh is} \textcolor[rgb]{0, 0.7, 0}{Republic of Artsakh}. It is located in \textcolor[rgb]{0, 0.7, 0}{Asia}. The capital of Republic of Artsakh is \textcolor[rgb]{0, 0.7, 0}{Stepanakert} ... \\
$\bullet$ (LT-1) \textbf{FiNE}: \textit{The name of the mother of Prince Philip, Duke of Edinburgh is} \textcolor[rgb]{0, 0.7, 0}{Princess Alice of Battenberg} ... \\
$\bullet$ (LT-2) \textbf{FiNE}: \textit{The gender of Prince Philip, Duke of Edinburgh is} \textcolor[rgb]{0, 0.7, 0}{male}. Prince Philip, Duke of Edinburgh is a member of the royal family. ... \\
\midrule
\multicolumn{1}{c}{\bf LLaMA-3} \\
\midrule
\vspace*{-7pt}
\textbf{ROME}: \textit{The name of the country of citizenship of Prince Philip, Duke of Edinburgh is} \textcolor[rgb]{0, 0.7, 0}{Republic of Artsakh (Nagorno-Karabakh Republic)} according to the Constitution and legislation of the Republic of Artsakh (Nagorno-Karabakh Republic) ... \\
$\bullet$ (LT-1) \textbf{ROME}: \textit{The name of the mother of Prince Philip, Duke of Edinburgh is} \textcolor[rgb]{0.9, 0, 0}{Svetlana Gogitidze} ... \\
$\bullet$ (LT-2) \textbf{ROME}: \textit{The gender of Prince Philip, Duke of Edinburgh is} \textcolor[rgb]{0.9, 0, 0}{a question that has been raised since the death of the former leader of the breakaway republic} ... \\
\cdashline{1-1}
\vspace*{-3.5pt}
\textbf{MEMIT}: \textit{The name of the country of citizenship of Prince Philip, Duke of Edinburgh is} \textcolor[rgb]{0, 0.7, 0}{Republic of Artsakh}. Republic of Artsakh is a state with limited recognition. It is located in the \textcolor[rgb]{0, 0.7, 0}{South Caucasus region of Eurasia} ... \\
$\bullet$ (LT-1) \textbf{MEMIT}: \textit{The name of the mother of Prince Philip, Duke of Edinburgh is} \textcolor[rgb]{0.9, 0, 0}{Natalia Ardzinba}. Natalia Ardzinba is a doctor, a specialist in obstetrics and gynecology ... \\
$\bullet$ (LT-2) \textbf{MEMIT}: \textit{The gender of Prince Philip, Duke of Edinburgh is} \textcolor[rgb]{0, 0.7, 0}{male}. \textcolor[rgb]{0.9, 0, 0}{Gender of Prince Philip, Duke of is male. Gender of Prince Philip, Duke of is Male. Gender of Prince Philip, Duke of is Male} ... \\
\cdashline{1-1}
\vspace*{-3.5pt}
\textbf{FiNE}: \textit{The name of the country of citizenship of Prince Philip, Duke of Edinburgh is} \textcolor[rgb]{0, 0.7, 0}{Republic of Artsakh}. The Republic of Artsakh is located in the \textcolor[rgb]{0, 0.7, 0}{Caucasus region of Eurasia} ... \\
$\bullet$ (LT-1) \textbf{FiNE}: \textit{The name of the mother of Prince Philip, Duke of Edinburgh is} \textcolor[rgb]{0, 0.7, 0}{Princess Alice of Battenberg}, and she is the \textcolor[rgb]{0, 0.7, 0}{sister of Louis Mountbatten} ... \\
$\bullet$ (LT-2) \textbf{FiNE}: \textit{The gender of Prince Philip, Duke of Edinburgh is} \textcolor[rgb]{0, 0.7, 0}{male}.
Prince Philip, Duke of Edinburgh was born on \textcolor[rgb]{0, 0.7, 0}{10th June 1921}. Philip was born in the Greek and Danish royal family ... \\
\bottomrule
\end{tabular}
\end{table}

\end{document}